\documentclass{article}

\usepackage{amsmath,amsfonts,bm}

\def\eqref#1{equation~\ref{#1}}

\def\1{\bm{1}}

\DeclareMathAlphabet{\mathsfit}{\encodingdefault}{\sfdefault}{m}{sl}
\SetMathAlphabet{\mathsfit}{bold}{\encodingdefault}{\sfdefault}{bx}{n}

\def\sR{{\mathbb{R}}}

\def\ov{{\boldsymbol{o}} }

\def\xv{\boldsymbol{x}}
\def\yv{\boldsymbol{y}}

\usepackage{microtype}
\usepackage{graphicx}
\usepackage{subfigure}
\usepackage{booktabs}
\usepackage{natbib}
\usepackage{algorithm}
\usepackage{algorithmic}
\usepackage{hhline}
\usepackage{multirow}
\usepackage{array}
\usepackage{tabularx}
\usepackage{makecell}
\usepackage{hyperref}

\usepackage{geometry}
\usepackage{amsmath}
\usepackage{amssymb}
\usepackage{mathtools}
\usepackage{amsthm}

\usepackage{color, colortbl}
\definecolor{Gray}{gray}{0.93}
\definecolor{orange}{rgb}{0.9,0.5,0}

\usepackage[capitalize,noabbrev]{cleveref}

\theoremstyle{plain}

\theoremstyle{definition}

\theoremstyle{remark}

\title{Mixing and Shifting: Exploiting Global and Local Dependencies in Vision MLPs}
\author{
Huangjie Zheng$^{1,2}$, ~Pengcheng He$^2$, ~Weizhu Chen$^2$, Mingyuan Zhou$^1$
\vspace{3mm}\\
\texttt{huangjie.zheng@utexas.edu, penhe@microsoft.com,}\\\texttt{ wzchen@microsoft.com, mingyuan.zhou@mccombs.utexas.edu}
}

\date{The University of Texas at Austin$^1 \qquad$  Microsoft Azure AI$^2$}
\begin{document}

\maketitle

\begin{abstract}
Token-mixing multi-layer perceptron (MLP) models have shown competitive performance in computer vision tasks with a simple architecture and relatively small computational cost. Their success in maintaining computation efficiency is mainly attributed to avoiding the use of self-attention that is often computationally heavy, yet this is at the expense of not being able to mix tokens both globally and locally. In this paper, to exploit both global and local dependencies without self-attention, we present Mix-Shift-MLP (MS-MLP) which makes the size of the local receptive field used for mixing increase in respect to the amount of spatial shifting. In addition to conventional mixing and shifting techniques, MS-MLP mixes both neighboring and distant tokens from fine- to coarse-grained levels and then gathers them via a shifting operation. This directly contributes to the interactions between global and local tokens. Being simple to implement, MS-MLP achieves competitive performance in multiple vision benchmarks. For example, an MS-MLP with 85 million parameters achieves $83.8\%$ top-1 classification accuracy on ImageNet-1K. Moreover, by combining MS-MLP with state-of-the-art Vision Transformers such as the Swin Transformer, we show MS-MLP achieves further improvements on  three different model scales, \textit{e.g.}, by $0.5\%$ on ImageNet-1K classification with Swin-B. The code is available at: \url{https://github.com/JegZheng/MS-MLP}.
\end{abstract}

\section{Introduction}
Showing promise in modeling  visual dependencies, Vision Transformers (ViTs) have advanced the state of the art (SoTA) of many different visual tasks \citep{vit, touvron2020training, liu2021Swin}. However, the self-attention module~\citep{vaswani2017attention}, which is key to the ViT success in capturing long-range visual dependencies, involves a computationally  intensive operation that  compares pairwise similarity between tokens. Inspired by self-attention, but without its heavy computation, several works show that building models solely on multi-layer perceptrons~(MLPs) can achieve surprisingly promising results on ImageNet~\citep{deng2009imagenet} classification with both spatial- and channel-wise token mixing~\citep{mlp-mixer, resmlp, gmlp}. These MLP-based models are efficient in token mixing to aggregate the spatial information and model visual feature dependencies, achieving results competitive to previous models on several representative computer vision tasks, such as image classification, object detection, and semantic segmentation.

Extensive studies of MLPs can be categorized into two mainstream directions depending on whether capturing global or local visual dependencies.   
Inspired by ViTs, global-mixing MLP-based methods such as MLP-Mixer~\citep{mlp-mixer} and ResMLP~\citep{resmlp} achieve the global reception field with the communication between patch tokens through spatial-wise projections. In this direction, researchers explore to effectively handle all tokens with various techniques, such as gating, routing, and Fourier transforms~\citep{gmlp,lou2021sparse,rao2021global,tang2021sparse,tang2021image}. Apart from MLPs that explore the modeling of global visual dependencies, a large number of studies have also achieved progress in using MLP-based architectures to model local visual dependencies, as done in the classical convolution paradigm~\citep{lecun1995convolutional}. Different from global-mixing architectures, local-mixing MLPs sample nearby tokens for interactions. In this direction several studies achieve effective token sampling by exploiting spatial shifting, permutation, and pseudo-kernel mixing~\citep{yu2021s,hou2021vision,mao2021rethinking,lian2021mlp,chen2021cyclemlp,guo2021hire}, \textit{etc}.

Despite the success from both perspectives, MLPs still avoid the self-attention at the expense of not being able to mix tokens as flexibly and efficiently as self-attention. The global-mixing is less flexible in identifying the importance among all tokens, while local-mixing is not able to capture long-range dependencies. In this paper, we investigate whether MLPs can effectively capture both short- and long-range dependencies to further improve performance. Intuitively, the visual dependencies between neighboring regions are usually more significant and need more attention, while those far away are still not trivial at a glance. Therefore, we propose to mix tokens from fine- to coarse-levels, where we perform fine-grained mixing in neighboring regions to achieve token interactions locally, while coarse-grained mixing for distant tokens to capture long-range dependencies. Specifically, we propose a multi-scale regional mixing, where the size of the regional receptive field used for mixing is proportional to the relative distance with respect to the query token, and these multi-scale regions are aggregated with a shifting operation. We plug in such multi-scale regional mixing into MLP architecture as Mix-Shift-MLP (MS-MLP). We evaluate the performance of MS-MLP via a comprehensive empirical study on a series of representative computer vision tasks, including image classification, object detection, and segmentation. According to the results, given a similar model complexity, our MS-MLP consistently outperforms SoTA MLPs across various settings, \textit{e.g.}, MS-MLP-B achieves 83.8\% in ImageNet-1K classification, which is on par with Focal-Attention-B~\citep{yang2021focal} with superior throughputs. In addition, we plug our MS-MLP module into SoTA Transformer models, which further improves both performance and efficiency. Notably, with MS-MLP, Swin Transformers \citep{liu2021Swin} and Focal Transformers~\citep{yang2021focal} respectively get improved on average by 0.5-0.6\% / 0.2-0.6\% in image classification and 0.2-0.5\% /  0.1-0.3\% in object detection and segmentation.

\section{Related works}
\paragraph{Global token-mixing MLPs:}
Global token-mixing MLPs are first proposed as self-attention-free alternatives to Transformer architectures~\citep{mlp-mixer,melas2021you,resmlp}. MLP-Mixer~\citep{mlp-mixer} replaces the self-attention layer of ViT with a spatial-wise MLP projection of tokens, achieving results that are competitive with ViT. gMLP \citep{gmlp}, consisting of an MLP-based module with multiplicative gating, provides competitive results in both vision and natural language processing (NLP) tasks. Vision Permutator \citep{hou2021vision} focuses on global mixing along both the vertical and horizontal axes. Raft-MLP~\citep{tatsunami2021raftmlp} employs a hierarchical and serialized structure which continuously improves accuracy. Similar to the parameterization of query and key pairs in ViTs, Wave-MLP~\citep{tang2021image} reweighs the importance of tokens with the amplitude and phase modules parameterized by two MLP projections. 

\paragraph{Local token-mixing MLPs:}
Local token-mixing MLPs focus more on the token interactions at local regions and hence share more similarities with convolutional neural networks (CNNs) than with Transformers. They have also been proved to achieve good performance on computer vision tasks. For example, a spatially shifted MLP (S$^2$-MLP)~\citep{yu2021s,yu2022s2} takes spatial shifts in four directions and mixes them in a channel-wise manner to gather information from neighboring tokens. Similar to S$^2$-MLP, an axial-shifted MLP (AS-MLP)~\citep{lian2021mlp} changes the spatial shifts in both the horizontal and vertical axes to gather local region information. CycleMLP~\citep{chen2021cyclemlp} takes pseudo-kernels and sample tokens from different spatial locations for mixing. ConvMLP~\citep{li2021convmlp} incorporates convolution layers and a pyramid structure to achieve local token mixing. Hire-MLP~\citep{guo2021hire} rearranges tokens across local regions to gain performance and computational efficiency. 

\paragraph{ViTs and CNNs:}
Transformers~\citep{vaswani2017attention}, originated from the NLP area, have recently been applied to visual tasks. In ViTs, the input is processed as patch tokens and then self-attention  is used to  aggregate  spatial information globally~\citep{vit}. \citet{touvron2020training} explore how to train ViTs efficiently with a distillation strategy. The use of CNNs has a long history for visual tasks, with extensive works conducted on improving the design to aggregate the features from local convolution and enlarge the receptive field \citep{lecun1998gradient,alexnet,vggnet,googlenet,resnet}. Recent works are proposed to marry the advantages of both global attention in ViTs and local attention in CNNs. PVT~\citep{pvt} exploits a pyramid structure for gathering spatial information on dense prediction tasks such as object detection. TNT~\citep{han2021transformer} takes small Transformer blocks to capture local information. Swin Transformer~\citep{liu2021Swin} proposes shifted window attention in order to aggregate spatial information from local regions. Focal-Transformer~\citep{yang2021focal} extends the use of local shifted windows to different scales to efficiently capture both short- and long-range visual dependencies. Incorporating recent findings in Transformers, ConvNeXT~\citep{liu2022convnet} enlarges the receptive fields with larger kernels to capture global dependencies. Related to these recent progress, our work aims to improve MLP-based architectures to efficiently capture both short-range and long-range visual dependencies.

\begin{figure*}[t]
    \centering
    \includegraphics[width=.88\textwidth]{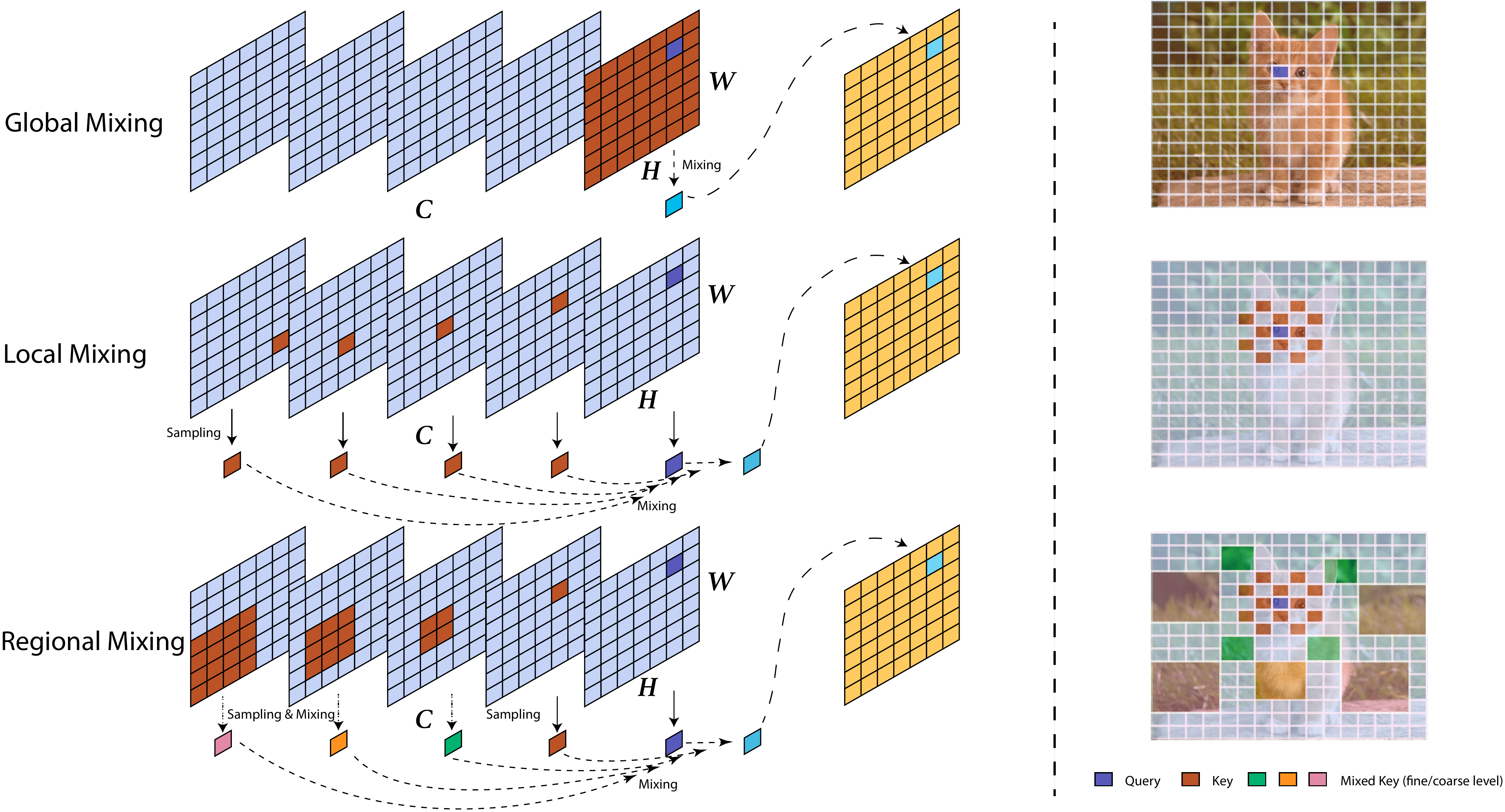}
    \caption{\small Illustrative comparison of typical operations in MLPs (best viewed in color). (\textbf{Left}) Different mixing strategies in a feature map: global mixing communicates within all the tokens to gather global spatial information; local mixing mechanism samples neighboring tokens to model local spatial dependency; regional mixing interacts tokens within regions of different scale that is proportional to the relative distance with regards to the query token. (\textbf{Right}) Corresponding visualization in view of attention mechanism: the global mixing captures dependency among all tokens, weighted by projection layer; the sparse mixing models local dependency with nearby tokens; our multi-scale regional mixing makes use of fine-grained level information for nearby tokens and coarse-level information for distant tokens. }
    \label{fig:comparison_mlp}
\end{figure*}

\section{Method}
Conventional MLPs are built upon a stacked architecture of multiple token-mixing blocks, where each token-mixing block consists of two sub-blocks, \textit{i.e.}, a spatial-mixing module and a channel-mixing MLP to aggregate spatial and channel information, respectively. Given an input feature with height $H$, width $W$, and channel $C$, expressed as $X\in \sR^{H\times W \times C}$, the token-mixing block is formulated as:
\begin{equation}
	\label{eq-tm-block}
	\begin{aligned}
		&Z= f_{\rm Spatial\text{-}Mixing} (h_{\rm Spatial} (X))+X, \\
		&O= f_{\rm Channel\text{-}Mixing} (h_{\rm Channel} (Z))+Z,
	\end{aligned}
\end{equation}
where $Z$ and $O$ denote the intermediate feature and output feature of the block, respectively, and $h$ denotes a normalization technique, such as batch normalization or layer normalization~\citep{bn,ln}. The channel-mixing function $f_{\rm Channel\text{-}Mixing}$ is usually parameterized with two MLPs, where the hidden dimension of the intermediate output is four times wider than the input dimension. Keeping this setting the same as previous ViTs and MLPs, we focus on investigating the spatial-mixing function $f_{\rm Spatial\text{-}Mixing}$ in what follows.

\subsection{Multi-scale regional token-mixing}\label{sec:intuition}
We first provide an illustration of the global and local token-mixing methods, as well as the proposed mixing that interacts tokens within regions on different scales. As shown in Fig.~\ref{fig:comparison_mlp}, given the query token patch (marked in cobalt blue), global mixing mixes all the tokens in the same channel together to get the output token (marked in cyan); local mixing first samples the tokens in nearby locations from all channels, then mixes them as the output. Correspondingly, in the right panel, the global mixing interacts with all locations and pays attention to all the other tokens, while the local mixing interacts with neighboring tokens of the query and pays attention to the nearby locations. Although global mixing gives interactions with all locations, it is more expensive in computation compared with local mixing, and none of the mixing schemes have the capability to give prioritization to the query. The receptive field of local mixing is also limited as the query only interacts with its neighbors. Here we consider a regional mixing with different region sizes, \textit{i.e.}, a fine-grained mixing in nearby locations and a coarse-grained mixing in distant locations for a global view. As shown in Fig.~\ref{fig:comparison_mlp}, we propose to sample in the closest regions at the finest granularity of the given query. As it samples more distant locations, we gradually increase the granularity and mix the tokens in that region to produce a coarse-grained token. Finally, similar to local mixing, we mix the tokens from channels ranging from the fine-grained tokens to coarse-grained tokens to produce the output token. As a result, region mixing has the ability to capture both global and local dependencies efficiently, where it covers many more regions in the feature map than local mixing and emphasizes more important local regions as well as capturing the distant regions with a coarse granularity when compared to global mixing.

\begin{figure*}[t]
    \centering
    \vspace{-1.5mm}
    \includegraphics[width=.85\textwidth]{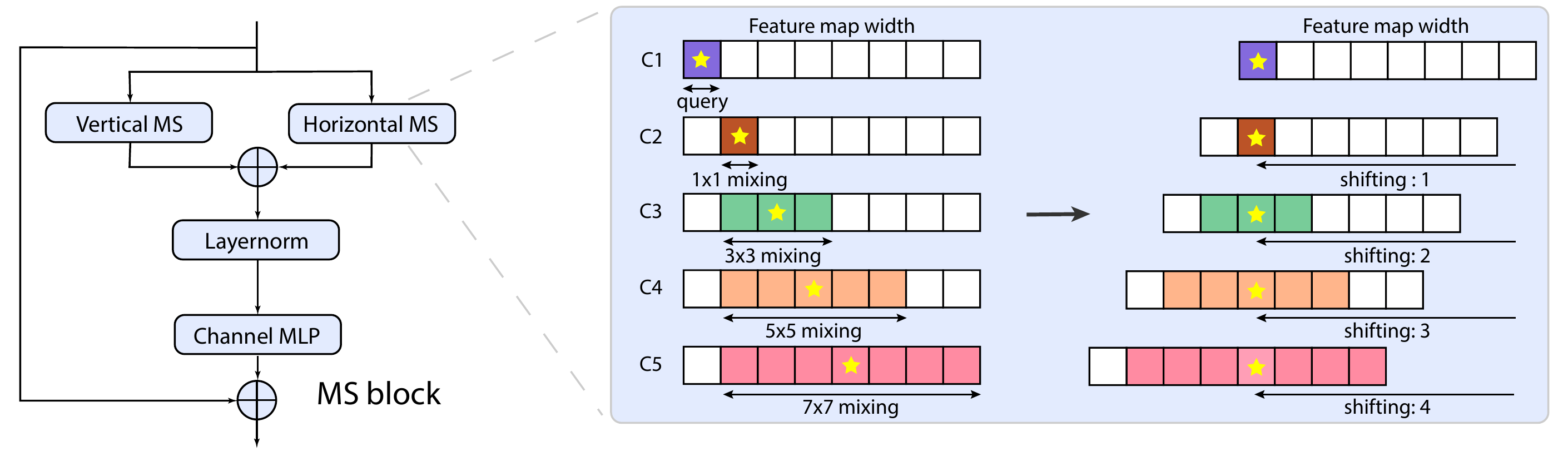}
    \vspace{-5mm}
    \caption{The plate notation of an MS-block ({\textbf{Left}}) and the illustration of the mixing-shifting operation in the horizontal direction of the feature map (\textbf{Right}). We first split the feature map into several groups (five groups here) along the channel dimension, with the first group regarded as the source of query tokens. In the other groups, as the centers of the attended regions (marked with yellow stars) become more and more distant, we gradually increase the mixing spatial range from $1\times 1$ to $7\times 7$. After the mixing operation, we shift the split channel groups to align their mixed center tokens with the query and then continue the channel-wise mixing with a channel MLP.}
    \label{fig:msblock}
    \vspace{-4mm}
\end{figure*}

Mathematically, with $\xv_c \in \sR^{H\times W}$ be the $c$-th channel in $X$, the global mixing can be formulated as the function $f_{\rm global}: \sR^{H\times W} \rightarrow \sR^{H \times W}$ such that $\ov_c= f_{\rm global} (\xv_c)$, where $f_{\rm global}$ is also parameterized with an MLP along the flattened $H \times W$ dimension \citep{mlp-mixer}. In this way, token $x_{i,j,c}$ gathers spatial information from tokens $\{x_{i^\prime, j^\prime}\}_{i^\prime \neq i, j^\prime \neq j}$ located elsewhere.  Supposing $\mathcal{N}_{ij} = \sR^{H^\prime \times W^\prime \times C} \subseteq \sR^{H \times W \times C}$ denotes the feature map neighboring location of center coordinate $(i,j)$, local mixing involves a spatial-wise sampling function $g_\mathrm{sampling}$ to get $\mathcal{N}_{ij}$. Then $f_\mathrm{local}(\,\cdot\, ;i,j): \mathcal{N}_{ij}(\cdot) \rightarrow \mathcal{N}_{ij}(\cdot)$ mixes the sampled token subset $\{x_{i^\prime, j^\prime}\}_{(i^\prime,   j^\prime) \in \mathcal{N}_{ij}}$ such that $\ov_{i,j} = f_\mathrm{local}(\{x_{i^\prime, j^\prime}\}_{(i^\prime,   j^\prime) \in \mathcal{N}_{ij}};{i,j})$, where $f_\mathrm{local}$ is usually a spatial arrangement operation like shifting or concatenation followed by a channel-wise projection \citep{yu2021s,lian2021mlp}. 

The proposed regional mixing combines the previous mixing strategies. We regard the sampled tokens $\{x_{i^\prime, j^\prime}\}_{(i^\prime,   j^\prime) \in \mathcal{N}_{ij}}$ as the center tokens of regions with different sizes and deploy a mixing function in each region as $f_{\rm region}^{r}$, where $f_{\rm region}^{r}: \sR^{H_r \times W_r} \rightarrow \sR^{H_r \times W_r}$, to let the center token represent information in that region, with $r$ denoting the size of the region. Below we describe this technique in detail.

\subsection{Regional token-mixing via mixing and shifting}\label{sec:method}
The regional token-mixing can be achieved with a composition of mixing and shifting, where they become dependent compared to conventional methods. Let's first define three terms to describe the proposed mixing-shifting clearly:

\textbf{ 1) Shifting size}: We denote $S$ as the shifting size of the feature map. We equally split the input feature along the channel into $S$ groups as $X = \xv_{C1} \cup ... \cup \xv_{CS}$ and $\forall n\neq m, \xv_{Cn} \cap \xv_{Cm} = \emptyset$. We assume the tokens in the first group $\xv_{C1}$ as queries and shift all the other groups.

\textbf{ 2) Relative distance}: We denote $d_n$ as the relative distance between the query group and each shifted groups $X_{Cn}$. We shift each feature group to reach a relative distance $d_n$ with regards to the query group.

\textbf{ 3) Mixing region size}: We denote $r_n$ as the mixing region size in $X_{Cn}$, where a larger $r_n$ indicates a coarser~granularity. Each group will take care of an $r_n \times r_n$ grid region.

Each time, like with local-mixing, we first target a center token $x_{i+d_n, j}$ or $x_{i, j+d_n}$ in each feature group $X_{Cn}$ and mix the tokens in an $r_n \times r_n$ grid region around the center token. Then, we shift the channel groups horizontally or vertically to both align the center tokens and get the features before channel-wise mixing:
\begin{equation}
	\label{eq-regional-mix}
	\begin{aligned}
		&\yv_{i+d_n, j,C_n}= f_{\rm region}^{r_n} (\xv_{i+d_n-r_n:i+d_n+r_n, j-r_n:j+r_n}), \\
		&\ov_{i,j}= f_{\rm local} (\{\yv_{i+d_n, j,C_n}\}_{n=1:S}),
	\end{aligned}
\end{equation}
where $f_{\rm region}^{r_n}$ is instantiated with a depth-wise convolution with kernel size $r_n \times r_n$ and $f_{\rm local}$ is a shifting operation.

Fig.~\ref{fig:msblock} shows the architecture of a mix-shifting block and an illustrative example of the proposed mix-shifting. In this example, we set $S=5$, $d_n = n - 1$, and $r_n$ = $2d_n - 1$. In this horizontal mix-shifting, the feature map is divided into five groups, with group C2-C4 gradually increasing both the region size and the relative distance between the center token and the query (cobalt blue token in C1). After the global mixing in each region, each group is shifted back according to the relative distance and all center tokens are aligned for the channel mixing.

\subsection{Complexity analysis}\label{sec:complexity}
In this subsection, we compare the computational complexity of typical ways to interact within tokens spatially, including the multi-head self-attention (MSA) in ViTs~\citep{vit}, window multi-head self-attention (W-MSA) in Swin Transformer~\citep{liu2021Swin}, focal multi-head self-attention (F-MSA) in Focal-Attention Transformer~\citep{yang2021focal}, global-mixing in MLP-Mixer~\citep{mlp-mixer}, local-mixing in AS-MLP~\citep{lian2021mlp}, and regional-mixing in our MS-MLP. We assume the channel-MLPs are the same, and that the shifting size and mixing region size in MS-MLP are the same as the focal level and focal region level in F-MSA. Denoting the input dimension as $H \times W \times C$ and the window size of W-MSA and F-MSA as $M$, the complexities of the above methods are shown as follows:
\begin{table}[ht]
  \centering
    \caption{Computation complexity comparison of different token interaction methods.} 
	\label{tab-complexity}
    \renewcommand{\arraystretch}{1.}
    \setlength{\tabcolsep}{1.0mm}{ 
	\setlength\tabcolsep{2pt}
    \resizebox{.8\columnwidth}{!}{%
    \begin{tabular}{l|c|c|c}
    \toprule[1.5pt]
    Method                                 & MSA & W-MSA & F-MSA \\ \hline 
    Complexity                             & $\mathcal{O}( 2(HW)^2C)$ & $\mathcal{O}( 2M^2HWC)$ & $\mathcal{O}( (S + \sum_{n=1}^S (r_n)^2) MHWC )$ \\ \hline \hline
    Method                                 & GM & AS & MS \\ \hline
    Complexity                             & $\mathcal{O}( (HW)^2C)$ & $\mathcal{O}( S )$ & $\mathcal{O}( \sum_{n=1}^S (r_n)^2)$ \\
    \bottomrule[1.5pt]
    \end{tabular}
    }} 
\end{table}

\begin{table*}[t]
\setlength{\tabcolsep}{2.1pt}
\footnotesize
    \centering
    \caption{MS-MLP model architectures with different configurations. Following the convention, we introduce  three different configurations---Tiny, Small, and Base---for different model capacities.}
    \label{tab:model_config}
    \resizebox{\linewidth}{!}{
    \begin{tabular}{l|c|c|c|c|c}
    \toprule[1.5pt]
            &  Input resolution & Layer Name & MS-MLP-Tiny & MS-MLP-Small & MS-MLP-Base \\
            \midrule
    \multirow{4}{*}{stage 1}  &   \multirow{4}{*}{$56\times56$}  & Patch Embedding  &  $p_1=4;c_1=96$ & $p_1=4;c_1=96$ & $p_1=4;c_1=128$ \\
    \cmidrule{3-6}
                              &   &  \makecell{MS-block} & \multicolumn{1}{c}{$\left[\begin{array}{c}
                                   S = 5 \\
                                   d_{1:S} = [0, 1, 2, 3, 4] \\
                                   r_{1:S} = [1, 1, 3, 5, 7]
                              \end{array}\right] \times 3$} &
\multicolumn{1}{c}{$\left[\begin{array}{c}
                                   S = 5 \\
                                   d_{1:S} = [0, 1, 2, 3, 4] \\
                                   r_{1:S} = [1, 1, 3, 5, 7]
                              \end{array}\right] \times 3$} &
\multicolumn{1}{c}{$\left[\begin{array}{c}
                                   S = 5 \\
                                   d_{1:S} = [0, 1, 2, 3, 4] \\
                                   r_{1:S} = [1, 1, 3, 5, 7]
                              \end{array}\right] \times 3$}
                              \\
        \midrule
    \multirow{4}{*}{stage 2} & \multirow{4}{*}{$ 28\times 28$} & Patch Embedding & $p_2=2;c_2=192$ & $p_2=2;c_2=192$ & $p_2=2;c_2=256$ \\
    \cmidrule{3-6}
            &  & \makecell{MS-block} & \multicolumn{1}{c}{$\left[\begin{array}{c}
                                    S = 5 \\
                                   d_{1:S} = [0, 1, 2, 3, 4] \\
                                   r_{1:S} = [1, 3, 3, 5, 7]
                              \end{array}\right] \times 3$} & 
\multicolumn{1}{c}{$\left[\begin{array}{c}
                                    S = 5 \\
                                   d_{1:S} = [0, 1, 2, 3, 4] \\
                                   r_{1:S} = [1, 3, 3, 5, 7]
                              \end{array}\right] \times 3$} &
\multicolumn{1}{c}{$\left[\begin{array}{c}
                                    S = 5 \\
                                   d_{1:S} = [0, 1, 2, 3, 4] \\
                                   r_{1:S} = [1, 3, 3, 5, 7]
                              \end{array}\right] \times 3$}                              \\
        \midrule
    \multirow{4}{*}{stage 3} & \multirow{4}{*}{$ 14\times 14$} & Patch Embedding & $p_3=2;c_3=384$ & $p_3=2;c_3=384$ & $p_3=2;c_3=512$\\
        \cmidrule{3-6}
            &  & \makecell{MS-block} & 
\multicolumn{1}{c}{$\left[\begin{array}{c}
                                    S = 5 \\
                                   d_{1:S} = [0, 1, 2, 3, 4] \\
                                   r_{1:S} = [1, 5, 5, 5, 7]
                              \end{array}\right] \times 9$} & \multicolumn{1}{c}{$\left[\begin{array}{c}
                                    S = 5 \\
                                   d_{1:S} = [0, 1, 2, 3, 4] \\
                                   r_{1:S} = [1, 5, 5, 5, 7]
                              \end{array}\right] \times 27$} & 
                 \multicolumn{1}{c}{$\left[\begin{array}{c}
                                    S = 5 \\
                                   d_{1:S} = [0, 1, 2, 3, 4] \\
                                   r_{1:S} = [1, 5, 5, 5, 7]
                              \end{array}\right] \times 27$}\\
        \midrule
    \multirow{4}{*}{stage 4} & \multirow{4}{*}{$ 7\times 7$ } & Patch Embedding & $p_4=2;c_4=768$ & $p_4=2;c_4=768$ & $p_4=2;c_4=1024$\\
    \cmidrule{3-6}
            &  & \makecell{MS-block} & \multicolumn{1}{c}{$\left[\begin{array}{c}
                                    S = 5 \\
                                   d_{1:S} = [0, 1, 2, 3, 4] \\
                                   r_{1:S} = [1, 7, 7, 7, 7]
                              \end{array}\right] \times 3$} &
\multicolumn{1}{c}{$\left[\begin{array}{c}
                                    S = 5 \\
                                   d_{1:S} = [0, 1, 2, 3, 4] \\
                                   r_{1:S} = [1, 7, 7, 7, 7]
                              \end{array}\right] \times 3$} &
\multicolumn{1}{c}{$\left[\begin{array}{c}
                                    S = 5 \\
                                   d_{1:S} = [0, 1, 2, 3, 4] \\
                                   r_{1:S} = [1, 7, 7, 7, 7]
                              \end{array}\right] \times 3$}                              \\
            \bottomrule[1.5pt]
    \end{tabular}}
\end{table*}

From Table~\ref{tab-complexity}, we can observe that compared with Transformers, MLPs largely reduce the computation complexity in dealing with token dependencies. In the comparison within MLPs, axial-shift and our proposed mix-shifting technique possess much lower complexity than global-mixing in MLP-Mixer. Note that here we make the shifting size $S$ and mixing region size $r_n$ the same as the focal level and focal region size in F-MSA, respectively, for better comparison, meaning each channel group has a different $r_n$. In practice, $r_n$ is not necessarily dependent on $S$, \textit{i.e.}, we can also set a focal level that is smaller than $S$ and the complexity is still $\mathcal{O}( \sum_{n=1}^S (r_n)^2)$. If we set the focal level to 1 and fix $r_n = 1$, this special case of MS-MLP will reduce to an AS-MLP~\citep{lian2021mlp}.

\subsection{Model architecture overview}\label{sec:architecture}
In this part we present an overview of the MS-MLP architecture. Following convention, we consider MS-MLP with Tiny, Small, and Base, corresponding to three different network configurations with different model capacities. Here we follow previous works to adopt a pyramid-like architecture~\citep{pvt,wu2021cvt,liu2021Swin,yang2021focal,guo2021hire} for our MS-MLP. Our model takes $224 \times 224$ pixel images as inputs and first splits the input image into patches (tokens) by a patch embedding. Then the token features go through a four-stage architecture. As the features go deeper to another stage, the number of tokens is reduced by a patch-embedding layer with ratio $p_i$ and output channels are simultaneously increased by this ratio. The spatial reduction ratio $p_i$ for these four stages is set as $[4,2,2,2]$. An overview of three configurations is shown in Table~\ref{tab:model_config}. We keep $S=5$ for all MS-blocks in the architecture for simplicity in presentation, though we believe there should exist a better configuration and we leave it to future exploration. Since the feature map resolution becomes smaller and smaller, we gradually decrease the fine-grained level region mixing. For example, at stage 4 we keep only the channel group with a $7\times 7$ region mixing, since in the last stage the patch resolution is sufficiently small. We find the architecture in Table~\ref{tab:model_config} performs the best; we also explored a simpler configuration in our ablations, where all MS-blocks keep the same configuration ($d_n = n - 1$ and $r_n$ = $2d_n - 1$) and the performance differed very slightly. Moreover, to match the number of  parameters and floating-point operations per second (FLOPs) of most existing models, we increase the number of MS-blocks in each stage, but keep the ratio of 1:1:3:1. 

\subsection{Improving Transformers in low-level stages}
How to combine the effectiveness of MLPs in computation with the flexibility of self-attention in Transformers is an interesting topic in computer vision research. In low-level stages, the model needs to process high-resolution inputs. Compared to W-MSA and local-mixing MLPs, MS-MLP has a much larger receptive field coverage; compared to MSA and global-mixing, MS-MLP covers just as many regions, but is more efficient with high-resolution inputs according to our analysis in Section~\ref{sec:complexity}. We empirically find MS-MLP can be combined with Transformers to boost performance, either by replacing the first-stage architecture or by being added as an additional stage to deal with finer-grained input (\textit{e.g.}, input with patch size 2). In our experiments, we show corresponding improvements over both the Swin Transformer~\citep{liu2021Swin} and Focal-Attention Transformer~\citep{yang2021focal}.

\begin{table}[t]
\centering
    \caption{\small Comparison of the proposed MS-MLP architecture with existing vision  MLP models on ImageNet. All models are trained and evaluated on $224 \times 224$ resolution, grouped according to the model size. Baseline results are quoted from the original papers.}
    \label{tab:comparison_MLP}
    \renewcommand{\arraystretch}{1.}
    \setlength{\tabcolsep}{1.0mm}{ 
    \resizebox{.85\columnwidth}{!}{%
    \begin{tabular}{l|c|ccc|c}
		\toprule[1.5pt]
		\multirow{2}{*}{Model} &  \multirow{2}{*}{Mixing.} &  \multirow{2}{*}{Params.} & \multirow{2}{*}{FLOPs} & Throughput &  Top-1 \\ 
		&&&&(image / s)&acc. (\%) \\ \hline
        gMLP-Ti~\citep{gmlp} &global & 6M & 1.4G & - & 72.3 \\
        ResMLP-S12~\citep{resmlp} &global & 15M & 3.0G & 1415.1 & 76.6 \\
        CycleMLP-B1~\citep{chen2021cyclemlp} &local & 15M & 2.1G & 1038.4 & 78.9 \\
        Wave-MLP-T~\citep{tang2021image} &global &17M  &2.4G &1208  & \textbf{80.6} \\
        Hire-MLP-T~\citep{guo2021hire} &local & 18M & 2.1G & 1561.7 & 79.7 \\ 
        \midrule
        gMLP-S~\citep{gmlp} &global & 20M  & 4.5G  &-& 79.6 \\
        ViP-Small/7~\citep{hou2021vision} &local & 25M & - & 719.0 & 81.5 \\
        AS-MLP-T~\citep{lian2021mlp} &local & 28M & 4.4G & 863.6 & 81.3 \\
        CycleMLP-B2~\citep{chen2021cyclemlp} &local & 27M & 3.9G & 640.6 & 81.6 \\
        Wave-MLP-S~\citep{tang2021image}  &global &30M  &4.5G & 720  &\textbf{82.6} \\
        Hire-MLP-S~\citep{guo2021hire} &local & 33M & 4.2G & 807.6 & 82.1 \\ 
        \rowcolor{Gray}
        MS-MLP-T  (ours)     & regional                       & 28M   & 4.9G  &792.0 &{82.1} \\
        \midrule
        Mixer-B/16~\citep{mlp-mixer} &global & 59M & 12.7G & - & 76.4 \\
        S$^2$-MLP-deep~\citep{yu2021s} &local & 51M & 10.5G & - & 80.7 \\ 
        ViP-Medium/7 \citep{hou2021vision} &local & 55M & - & 418.0 & 82.7 \\
        CycleMLP-B4 \citep{chen2021cyclemlp} &local  & 52M & 10.1G & 320.8 & 83.0 \\
        AS-MLP-S~\citep{lian2021mlp} &local & 50M & 8.5G & 478.4 & 83.1 \\
        Wave-MLP-M~\citep{tang2021image} &global &44M  &7.9G  &413  &\textbf{83.4}  \\
        Hire-MLP-B~\citep{guo2021hire} &local & 58M & 8.1G & 440.6 & 83.2 \\ 
        \rowcolor{Gray}
        MS-MLP-S (ours)  & regional & 50M   & 9.0G  &483.8 & \textbf{83.4}\\
        \midrule	
        ResMLP-B24~\citep{resmlp} &global & 116M & 23.0G & 231.3 & 81.0 \\
        S$^2$-MLP-wide~\citep{yu2021s} &local & 71M & 14.0G & - & 80.0 \\ 
        CycleMLP-B5~\citep{chen2021cyclemlp} &local & 76M & 12.3G & 246.9 & 83.2 \\
        gMLP-B~\citep{gmlp}  &global & 73M & 15.8G & - & 81.6 \\
        ViP-Large/7~\citep{hou2021vision} &local & 88M & - & 298.0 & 83.2 \\
        AS-MLP-B~\citep{lian2021mlp} &local & 88M & 15.2G & 312.4 & 83.3 \\
        Wave-MLP-B~\citep{tang2021image} &global &63M  &10.2G  &341  &83.6 \\
        Hire-MLP-L~\citep{guo2021hire} &local & 96M & 13.4G & 290.1 & \textbf{83.8} \\ 
        \rowcolor{Gray}
        MS-MLP-B      (ours)    & regional                         & 88M   & 16.1G  &366.5 & \textbf{83.8}\\ 
		\bottomrule[1.5pt]
	\end{tabular}}
    }
\end{table}

\section{Experiments}
In this section, we investigate the effectiveness of the MS-MLP architectures using experiments on multiple vision tasks. We first use image classification on ImageNet-1K~\citep{deng2009imagenet} to compare MS-MLP with previous state-of-the-art MLPs. Next, we show the performance of combining MS-MLP with the Swin Transformer~\citep{liu2021Swin} and with the Focal-Attention Transformer~\citep{yang2021focal}, with a comparison to all SoTA methods on this task. Furthermore, we compare MS-MLP with existing alternatives using object detection and semantic segmentation on COCO-2017~\citep{coco}. Finally, we present ablation studies on the effects of regional mixing. Appendix~\ref{app:exp-settings} provides detailed experimental settings.

\subsection{Image classification on ImageNet-1K}
On ImageNet-1K~\citep{deng2009imagenet}, for a fair comparison, we follow the commonly used training recipes in~\citet{vit} and \citet{pvt}. All models are trained for 300 epochs with a batch size of 1,024. The initial learning rate is set to $10^{-3}$ with 20 epochs of linear warm-up starting from $10^{-5}$. For optimization, we use AdamW~\citep{adamw} as the optimizer with a cosine learning rate scheduler. The weight decay is set to $0.05$ and the maximal gradient norm is clipped to 5.0. We use the same set of data augmentation and regularization strategies as in~\citet{touvron2020training}, including Rand-Augment~\citep{randaugment}, MixUp~\citep{mixup}, CutMix~\citep{cutmix}, Label Smoothing~\citep{labelsmooth}, Random Erasing~\citep{randomerasing}, and DropPath~\citep{droppath}.
The stochastic depth drop rates are set to $0.2$, $0.3$, and $0.5$ for our Tiny, Small, and Base models, respectively. During training, we crop images randomly to $224 \times 224$, while a center crop is used during evaluation on the validation set. All models are trained on a node with eight NVIDIA Tesla V100 GPUs, based on which we report the experimental results with top-1 accuracy, number of parameters, FLOPs, and throughput.

\textbf{Main results comparing with MLPs}: We compare the proposed MS-MLP with previous MLP-based models on ImageNet, as shown in Table~\ref{tab:comparison_MLP}. MS-MLP with region mixing consistently achieves competitive results with better computation efficiency. For example, compared with AS-MLP~\citep{lian2021mlp} and CycleMLP~\citep{chen2021cyclemlp}, MS-MLP can perform significantly better with comparable parameter scales, FLOPs, and throughput. When compared with recently proposed Wave-MLP~\citep{tang2021image} and Hire-MLP~\citep{guo2021hire}, MS-MLP obtains a better throughput and similar  classification accuracy. In particular, scale up to Base configuration, MS-MLP achieves the best results (83.8\%) with a throughput (366.5 images/sec) surpassing all the other models with a comparable number of parameters. 

\begin{table}[t]
  \centering 
    \caption{Comparison of MS-MLP architecture with representative SoTA models on ImageNet-1K with a resolution of $224 \times 224$. }
	\label{tab-sota}
    \renewcommand{\arraystretch}{1.2}
    \setlength{\tabcolsep}{1.0mm}{ 
	\setlength\tabcolsep{2pt}
    \resizebox{.8\columnwidth}{!}{
    \begin{tabular}{l | c |c c c|l}
	\toprule[1.5pt]	
	
	\multirow{2}{*}{Model} & \multirow{2}{*}{Family}  & \multirow{2}{*}{Params.} & \multirow{2}{*}{FLOPs} & Throughput &  Top-1 \\ 
	&&&&(images / s)&acc. (\%)\\ \hline
	ResNet18~\citep{he2016deep}                   & CNN    & 12M & 1.8G  &-& 69.8 \\
	ResNet50~\citep{he2016deep}                   &  CNN   & 26M & 4.1G  &- &78.5 \\
	ResNet101~\citep{he2016deep}                  &  CNN  & 45M & 7.9G &-& 79.8 \\
	ConvNeXt-T~\citep{liu2022convnet}  & CNN & 29M & 4.5G & 775 & 82.1 \\
	ConvNeXt-S~\citep{liu2022convnet}  & CNN & 50M & 8.7G & 447 & 83.1 \\
	ConvNeXt-B~\citep{liu2022convnet}  & CNN & 89M & 15.4G & 292 & \textbf{83.8} \\
	\hline 
	Swin-T~\citep{liu2021Swin}                    & Trans &  29M & 4.5G  &755& 81.3 \\
	Swin-S~\citep{liu2021Swin}                    & Trans & 50M & 8.7G  & 437&83.0 \\ 
	Swin-B~\citep{liu2021Swin}                    & Trans &  88M & 15.4G &278 &83.3 \\  
	Focal-Attention-T~\citep{yang2021focal}                    & Trans &  29M & 4.9G  &319& 82.2 \\
	Focal-Attention-S~\citep{yang2021focal}                    & Trans & 52M & 9.4G  & 192&83.5 \\
	Focal-Attention-B~\citep{yang2021focal}                    & Trans &  90M & 16.4G &138 &\textbf{83.8} \\ \hline
	\rowcolor{Gray}
    MS-MLP-T  (ours)     & MLP                       & 28M   & 4.9G  &792 &{82.1} \\
	\rowcolor{Gray}
    MS-MLP-S (ours)  & MLP & 50M   & 9.0G  &484 & {83.4}\\
	\rowcolor{Gray}
    MS-MLP-B      (ours)    &  MLP                         & 88M   & 16.1G  &366 & \textbf{83.8}\\ 
	\bottomrule[1.5pt]
\end{tabular}}
    }
\end{table}

\begin{table}[t]
  \centering
    \caption{Comparison of Swin and Focal-Attention transformer w/o MS-MLP on ImageNet-1K. }
	\label{tab-trans}
    \renewcommand{\arraystretch}{1.2}
    \setlength{\tabcolsep}{1.0mm}{ 
	\setlength\tabcolsep{2pt}
    \resizebox{.8\columnwidth}{!}{
    \begin{tabular}{l | c |c c c|l}
	\toprule[1.5pt]	
	
	\multirow{2}{*}{Model} & \multirow{2}{*}{Family}  & \multirow{2}{*}{Params.} & \multirow{2}{*}{FLOPs} & Throughput &  Top-1 \\ 
	&&&&(images / s)&acc. (\%)\\ \hline
	Swin-T~\citep{liu2021Swin}                    & Trans &  29M & 4.5G  &755& 81.3 \\
	Swin-S~\citep{liu2021Swin}                    & Trans & 50M & 8.7G  & 437&83.0 \\ 
	Swin-B~\citep{liu2021Swin}                    & Trans &  88M & 15.4G &278 &83.3 \\  
	\rowcolor{Gray}
	MS-MLP + Swin-T   (ours)                  & MLP + T &  29M & 4.5G  &779 & \textbf{81.9}\\ 
	\rowcolor{Gray}
	MS-MLP  + Swin-S  (ours)                   & MLP + T & 50M & 8.7G  & 464 & \textbf{83.5}\\ 
	\rowcolor{Gray}
	MS-MLP  + Swin-B  (ours)                   & MLP + T &  88M & 15.4G &279 &\textbf{83.8}\\ \hline 
	Focal-Attention-T~\citep{yang2021focal}                    & Trans &  29M & 4.9G  &319& 82.2 \\
	Focal-Attention-S~\citep{yang2021focal}                    & Trans & 52M & 9.4G  & 192&83.5 \\
	Focal-Attention-B~\citep{yang2021focal}                    & Trans &  90M & 16.4G &138 &83.8 \\ 
	\rowcolor{Gray}
	MS-MLP  + Focal-Attention-T    (ours)                 & MLP + T              &  29M & 5.6G  &451& \textbf{82.8}\\ 
	\rowcolor{Gray}
	MS-MLP  + Focal-Attention-S    (ours)             & MLP + T                 & 52M & 10.1G  & 297 &\textbf{83.9}\\ 	
	\rowcolor{Gray}
	MS-MLP  + Focal-Attention-B    (ours)             & MLP + T                 & 90M & 17.6G & 207 &\textbf{84.0}\\ 
	\bottomrule[1.5pt]
\end{tabular}}
    }
\end{table}

\begin{table*}[t]
\centering
\caption{\small Object detection and instance segmentation results on COCO val2017. We compare MS-MLP with other backbones based on RetinaNet and Mask R-CNN frameworks. All models are trained on the ``1x" schedule. }
\label{table:coco-1x}
\renewcommand{\arraystretch}{0.88}
\setlength\tabcolsep{2.8pt}
\resizebox{\textwidth}{!}{\begin{tabular}{l|c|c|ccc|c|ccc|ccc}
\toprule[1.5pt]
\multirow{2}{*}{Backbone} &\multicolumn{5}{c|}{RetinaNet 1$\times$} &\multicolumn{7}{c}{Mask R-CNN 1$\times$} \\
\cline{2-13} 
& Param / FLOPs & AP & AP$_\mathrm{S}$ &AP$_\mathrm{M}$ & AP$_\mathrm{L}$ & Param / FLOPs & AP$^{\rm b}$ & AP$_{50}^{\rm b}$ &AP$_{75}^{\rm b}$  &AP$^{\rm m}$ &AP$_{50}^{\rm m}$ & AP$_{75}^{\rm m}$\\
\midrule

CycleMLP-B2~\citep{chen2021cyclemlp} & 36.5M / 230G & 40.9 & 23.4 & 44.7 & 53.4 & 46.5M / 249G & 41.7 & 63.6 & 45.8 & 38.2 & 60.4 & 41.0 \\
Wave-MLP-S~\citep{tang2021image} &37.1M / 231G & \textbf{43.4}  & \textbf{26.6} & 47.1 & 57.1 &47.0M / 250G & 44.0 & 65.8 &48.2 & 40.0 & \textbf{63.1} & 42.9\\
Hire-MLP-S~\citep{guo2021hire} & 42.8M / 237G & {41.7} & {25.3} & {45.4} & 54.6 & 52.7M / 256G & {42.8} & {65.0} & {46.7} & {39.3} & {62.0} & {42.1} \\
\rowcolor{Gray} 
MS-MLP-T (ours) & 39.6M / 265G & {42.7} & {26.0} & \textbf{{47.3}} & \textbf{59.7} & 49.8M / 269G & \textbf{44.4} & \textbf{67.8} & \textbf{47.8} & \textbf{40.4} & 61.3 & \textbf{44.2} \\ \hline
CycleMLP-B3~\citep{chen2021cyclemlp} & 48.1M / 291G & 42.5 & 25.2 & 45.5 & 56.2 & 58.0M / 309G & 43.4 & 65.0 & 47.7 & 39.5 & 62.0 & 42.4 \\
CycleMLP-B4~\citep{chen2021cyclemlp} & 61.5M / 356G & 43.2 & 26.6 & 46.5 & 57.4 & 71.5M / 375G & 44.1 & 65.7 & 48.1 & 40.2 & 62.7 & 43.5 \\
Wave-MLP-M~\citep{tang2021image} &49.4M / 291G & 44.8 & 28.0 & 48.2 & 59.1 & 59.6M / 311G & 45.3 & 67.0 & 49.5  & 41.0  &64.1 & 44.1 \\
Hire-MLP-B~\citep{guo2021hire} & 68.0M / 316G & 44.3 & {28.0} & {48.4} & 58.0 & 77.8M / 334G & {45.2} & {66.9} & {49.3} & {41.0} & {64.0} & {44.2} \\
\rowcolor{Gray}			    				
MS-MLP-S (ours) & 61.2M / 360G & \textbf{{45.0}} & \textbf{{28.2}} & \textbf{{48.9}} & \textbf{60.4} & 70.9M / 372G & \textbf{47.1} & \textbf{71.1} & \textbf{51.6} & \textbf{41.9} & \textbf{64.1} & \textbf{45.1} \\ \hline
CycleMLP-B5~\citep{chen2021cyclemlp} & 85.9M / 402G & 42.7 & 24.1 & 46.3 & 57.4 & 95.3M / 421G & 44.1 & 65.5 & 48.4 & 40.1 & 62.8 & 43.0 \\
Wave-MLP-B~\citep{tang2021image} & 66.1M / 334G & 44.2 & 27.1 & 47.8 & 58.9 & 75.1M / 353G & 45.7 & 67.5 & 50.1 & 27.8 & 49.2 & 59.7 \\
Hire-MLP-L~\citep{guo2021hire} & 105.8M / 424G & {44.9} & \textbf{28.9} & {48.9} & {57.5} & 115.2M / 443G & {45.9} & {67.2} & {50.4} & {41.7} & \textbf{64.7} & {45.3} \\
\rowcolor{Gray}					
MS-MLP-B (ours) & 97.3M / 544G  & \textbf{45.7} & 27.8 & \textbf{49.2} & \textbf{59.7} & 107.5M / 557G & \textbf{46.4} & \textbf{67.2} & \textbf{50.7} & \textbf{42.4} & {63.6} & \textbf{46.4} \\ 
\bottomrule[1.5pt]
\end{tabular}}
\end{table*}

\textbf{Comparing with SoTAs}: 
Besides MLPs, we compare MS-MLP with representative CNN-based and Transformer-based SoTA models. The input image resolution  is set as 224 $\times$ 224. As in Table~\ref{tab-sota}, compared with SoTAs, MS-MLP achieves competitive performance with better efficiency. For example, MS-MLP-B achieves 83.8\% top-1 accuracy, which is superior to Swin-B with 83.3\% accuracy. The computational efficiency of MS-MLP is significantly better than Transformers, and slightly surpasses the CNN architectures like ConvNeXt~\citep{liu2022convnet}.

\textbf{Results with Transformers}: 
To effectively combine the strengths of both MS-MLP and Transformers, we let MS-MLP and Transformers represent low-level and high-level stages, respectively. We found this design can largely boost model efficiency. In the experiments, we replace the first stage of the Swin Transformer~\citep{liu2021Swin} and the Focal-Attention Transformer~\citep{yang2021focal} with MS-blocks. Moreover, we add a stage zero that consists of two MS-blocks with $p_0 = 2, c_0 = c_1/2$ ahead of stage one. This novel configuration produces a modified model having similar parameter sizes, FLOPs, and throughputs as the original models, but a consistently better performance in terms of  accuracy.

We compare the proposed MS-MLP+Transformer architecture with the original architectures on ImageNet-1K and summarize the results in Table~\ref{tab-trans}. Compared with the Swin Transformer and Focal-Attention Transformer, the MS-MLP+Transformer architectures achieve both a higher accuracy and a higher throughput with a similar number of parameters and FLOPs. For example, with 88M parameters and 15.4G FLOPs, MS-MLP+Swin-B achieves an 83.8 top-1 accuracy, surpassing Swin-B with 83.3\% accuracy. With very similar 90M parameters and slightly higher FLOPs, MS-MLP+Focal-Attention improves both the top-1 classification accuracy and the throughputs, \textit{e.g.} Focal-B is improved to 84.0\%, which sets a new SoTA accuracy with a similar model size. The superiority of MS-MLP+Transformer clearly implies that the MS-MLP architecture has a better efficiency in the token aggregation process. This positive effect becomes more pronounced with larger token numbers, since the proposed mixing and shifting operations can exploit both global and local dependencies adequately.

\subsection{Object detection and instance segmentation}
\textbf{Results on COCO-2017}: 
We first conduct the object detection and instance segmentation experiments on COCO-2017 ~\citep{coco}. Following previous works~\citep{pvt,liu2021Swin}, we use the pretrained models as backbones and plug into RetinaNet~\citep{retinanet}, Mask R-CNN~\citep{maskrcnn}, and Cascade Mask R-CNN~\citep{cascadercnn} in mmdetection~\citep{mmdetection}. We adopt the single-scale and multi-scale training for the ``1x" and ``3x" schedules, respectively.

\begin{table}[t]
\centering
\caption{Results of semantic segmentation on the ADE20K validation set. FLOPs are calculated with an input size of 2048$\times$512.}
\label{tab:ade20k}
\renewcommand{\arraystretch}{0.88}
\setlength\tabcolsep{4.4pt}
\resizebox{.65\columnwidth}{!}{
\begin{tabular}{l|c|c|c|c}
\toprule[1.5pt]
Backbone & Param & FLOPs & SS mIoU & MS mIoU \\
\midrule  
 Swin-T~\citep{liu2021Swin} & 60M & 945G  & 44.5 & 46.1 \\
 Focal-T~\citep{yang2021focal} &62M & 998G & 45.8 & 47.0 \\
 AS-MLP-T~\citep{lian2021mlp} & 60M & 937G  & - & 46.5 \\
 Hire-MLP-S~\citep{guo2021hire}& 63M & 930G  & \textbf{46.1} & \textbf{47.1} \\
 \rowcolor{Gray}
 MS-MLP-T (ours) & 61M & 939G  & {46.0} & {46.8} \\
\midrule
ResNet-101~\citep{resnet} & 86M & 1029G  & 43.8 & 44.9 \\
Swin-S~\citep{liu2021Swin} & 81M & 1038G  & 47.6 & 49.5 \\
Focal-S~\citep{yang2021focal} &85M & 1130G & 48.0 & \textbf{50.0} \\
AS-MLP-S~\citep{lian2021mlp} & 81M & 1024G  & - & 49.2 \\
Hire-MLP-B~\citep{guo2021hire} & 88M & 1011G  & {48.3} & {49.6} \\
\rowcolor{Gray}
 MS-MLP-S (ours) & 82M & 1028G  & \textbf{48.7} &{49.6} \\
\midrule
Swin-B~\citep{liu2021Swin} & 121M & 1188G  & 48.1 & 49.7 \\
Focal-B~\citep{yang2021focal} &126M & 1354G & 49.0 & \textbf{50.5} \\
AS-MLP-B~\citep{lian2021mlp} & 121M & 1166G  & - & 49.5 \\
Hire-MLP-L~\citep{guo2021hire} & 127M & 1125G  & 48.8 & {49.9} \\
\rowcolor{Gray}
 MS-MLP-B (ours) & 122M & 1172G  & \textbf{49.1} & {49.9} \\
\bottomrule[1.5pt]
\end{tabular}
}
\end{table}

\begin{table}[t]
\centering
\vspace{-2.5mm}
\caption{\small Analogous comparison to Table~\ref{table:coco-1x} between using MS-MLP plugged-in Transformer architecture and their original architectures. }
\label{table:coco-mlp-trans-1x}
\renewcommand{\arraystretch}{0.88}
\setlength\tabcolsep{2.8pt}
\resizebox{\textwidth}{!}{\begin{tabular}{l|c|c|ccc|c|ccc|ccc}
\toprule[1.5pt]
\multirow{2}{*}{Backbone} &\multicolumn{5}{c|}{RetinaNet 1$\times$} &\multicolumn{7}{c}{Mask R-CNN 1$\times$} \\
\cline{2-13} 
& Param / FLOPs & AP & AP$_\mathrm{S}$ &AP$_\mathrm{M}$ & AP$_\mathrm{L}$ & Param / FLOPs & AP$^{\rm b}$ & AP$_{50}^{\rm b}$ &AP$_{75}^{\rm b}$  &AP$^{\rm m}$ &AP$_{50}^{\rm m}$ & AP$_{75}^{\rm m}$\\
\midrule
Swin-T~\citep{liu2021Swin} & 38.5M / 244G & 41.5 & \textbf{25.1} & 44.9 & {55.5} & 47.8M / 264G & 42.2 & 64.6 & \textbf{46.2} & 39.1 & \textbf{61.6} & 42.0 \\
\rowcolor{Gray} 
MS-MLP+Swin-T (ours) & 38.5M / 262G & \textbf{{42.0}} & 24.5	 &\textbf{45.4}	&\textbf{57.4} & 49.8M / 269G & \textbf{42.7}	& \textbf{64.8} & 45.9 & \textbf{39.5} & 59.8 & \textbf{42.3} \\ \hline
Focal-Attention-T~\citep{yang2021focal} & 39.4M / 265G  & 43.7 & \textbf{28.6} & \textbf{47.4} & 56.9 & 44.8M / 291G & \textbf{{44.8}} & \textbf{{67.7}} &\textbf{{49.2}} &\textbf{ {41.0}} & \textbf{{64.7}} & {44.2} \\ 
\rowcolor{Gray} 
MS-MLP+Focal-Attention-T (ours) & 39.4M / 284G & \textbf{{44.3}} & {27.0} & {47.3} & \textbf{58.9} & 44.8M / 299G & 44.6 & 65.7 & 48.7 & 40.6 & 62.6 & \textbf{44.7} \\
\hline \midrule		
Swin-S~\citep{liu2021Swin} & 59.8M / 334G & {44.5} & 27.4 & 48.0 & \textbf{59.9} & 69.1M / 353G & 44.8 & 66.6 & 48.9 & 40.9 & 63.4 & {44.2} \\
\rowcolor{Gray}						
MS-MLP+Swin-S (ours) & 59.9M / 344G  & \textbf{45.2} & \textbf{27.9} & \textbf{48.4} & {59.5} & 69.1M / 379G & \textbf{45.3} & \textbf{67.2} & \textbf{49.6} & \textbf{41.0} & \textbf{64.1} & \textbf{ 44.1} \\ \hline
Focal-Attention-S~\citep{yang2021focal} & 61.7M / 367G  & {45.6} & \textbf{29.5} & {49.5} & {60.3} & 71.2M / 401G & {47.4} & \textbf{69.8} & {51.9} & {42.8} & {66.6} & {46.1} \\
\rowcolor{Gray} 		
MS-MLP+Focal-Attention-S (ours) & 62.0M / 354G  & \textbf{45.8} & {27.8} & \textbf{51.0} & \textbf{60.5} & 72.7M / 446G & \textbf{47.5} & {66.5} & \textbf{52.3} & \textbf{43.0} & \textbf{67.3} & \textbf{46.4} \\
\hline \midrule	
Swin-B~\citep{liu2021Swin} & 98.4M / 477G & 45.0 & \textbf{28.4} & 49.1 & \textbf{60.6} & 107M / 496G  & 46.9 & 69.2 & \textbf{51.6} & 42.3 & 66.0 & 45.5 \\
\rowcolor{Gray}							
MS-MLP+Swin-B (ours) & 98.7M / 502G & \textbf{45.3} & 28.0 & \textbf{49.4} & 59.5 & 108.6M / 561G  & \textbf{47.2} & \textbf{69.5} & 51.3 & \textbf{42.8} & 66.0 & \textbf{46.3} \\ \hline
Focal-Attention-B~\citep{yang2021focal} & 100.8M / 514G  & {46.3} & \textbf{31.7} & \textbf{50.4} & {60.8} & 110.0M / 533G & {47.8} & {70.2} & \textbf{52.5} & {43.2} & {67.3} & {46.5} \\ 
\rowcolor{Gray}				
MS-MLP+Focal-Attention-B (ours) & 104.2M / 609G  & \textbf{46.6} & {28.8} & \textbf{50.4} & \textbf{62.2} & 116.9M / 647G & \textbf{47.9} & \textbf{70.6} & 52.4 & \textbf{43.4} & \textbf{67.8} & \textbf{47.8} \\
\bottomrule[1.5pt]
\end{tabular}}
\end{table}

The results of object detection and instance segmentation with different frameworks and training schedules are reported in Table~\ref{table:coco-1x} and Table~\ref{table:coco-3x} (Appendix), respectively. As shown in Table~\ref{table:coco-1x}, using MS-MLP as the backbone, RetinaNet and Mask R-CNN surpass most of the MLP-based baselines. For example, compared to recent MLPs, MS-MLP-B outperforms Hire-MLP-L and WaveMLP-B by 0.6\% and 1.5\%, respectively, on RetinaNet. In downstream tasks like object detection and segmentation, Hire-MLP-L and Wave-MLP-B usually require high-resolution input and both short- and long-range token interactions. Our regional mixing shows the effectiveness from this perspective, compared to the global-only or local-only token interactions.

\textbf{Results on ADE20K}: 
Besides the detection and instance segmentation tasks, we further evaluate our model on semantic segmentation, where we use UperNet~\citep{upernet} with our pretrained models as the backbone.  For all models, we use a standard recipe by setting the input size as $512 \times 512$ and train the model for 160k iterations with a batch size of 16. The results are shown in Table~\ref{tab:ade20k}. We can observe that MS-MLP achieves better single-scale mIoUs than the baselines and has competitive multi-scale mIoUs across different model capacities.

\textbf{Results with Transformers}: 
Similar to the classification tasks, we validate the effectiveness of combining MS-MLP with the Swin and Focal-Attention Transformers on COCO-2017 object detection and instance segmentation using their original settings. Table~\ref{table:coco-mlp-trans-1x} shows an analogous comparison to Table~\ref{table:coco-1x} for this task. The MS-MLP+Transformer backbone improves their original Transformer backbones by 0.5-0.7\% on both RetinaNet and Mask-RCNN.

\begin{table}[t]
\begin{minipage}[t]{.62\textwidth}
  \centering
    \caption{Configurations of different $r_n$ and $d_n$ of MS-MLP.} 
	\label{tab-ablation}
    \renewcommand{\arraystretch}{1.3}
    \setlength{\tabcolsep}{1.0mm}{ 
	\setlength\tabcolsep{2pt}
    \resizebox{\columnwidth}{!}{\begin{tabular}{l|c|c|c|c}
    \toprule[1.5pt]
    Configs & Local & Global & Isolated Regional & Regional\\ \hline
    $r_{1:S}$& $[1,1,1,1,1]$ & $[7,7,7,7,7]$& $[1,1,3,5,7]$ & $[1,1,3,5,7]$ \\ \hline 
    $d_{1:S}$& $[0,1,2,3,4]$ & $[0,1,2,3,4]$& $[0,2,5,10,17]$ & $[0,1,2,3,4]$ \\
    \bottomrule[1.5pt]
    \end{tabular}
    }
    }
\end{minipage}\hfill
\begin{minipage}[t]{.34\textwidth}
\centering
    \caption{\small Comparison of different patch size on ImageNet-1K.} 
	\label{tab-patch}
    \renewcommand{\arraystretch}{.7}
    \setlength{\tabcolsep}{.8mm}{ 
	\setlength\tabcolsep{0.8pt}
    \resizebox{\columnwidth}{!}{\begin{tabular}{l|c|c}
    \toprule[1.5pt]
    {\small Method}                                 & {\small Patch size} & {\small Accuracy} \\ \hline
    \multirow{2}{*}{\small Swin-T}             & {\small 2}          & {\small 81.4\%}  \\
                                           & {\small 4}          & {\small 81.3\%}  \\\hline
    \multirow{2}{*}{\small MS-MLP+Swin-T}          & {\small 2}          & {\small 81.9\%}  \\
                                           & {\small 4}          & {\small 81.6\%}  \\
    \bottomrule[1.5pt]
    \end{tabular}
    }}
\end{minipage}
\end{table}

\subsection{Ablation studies}
We conduct ablation studies using MS-MLP-Tiny by varying the region mixing size $r_n$ and relative distance $d_n$ to validate the effectiveness of regional mixing. We propose four different configurations as shown in Table~\ref{tab-ablation}: 1) all regional mixing sizes are restricted to 1 to achieve local mixing. This keeps only the shifting module and is close to AS-MLP~\citep{lian2021mlp}. 2) All region sizes are set to 7 to cover as many tokens as possible. This setting aims to approximate the global mixing and is also related to W-MSA in Swin~\citep{liu2021Swin}. 3) We set the region to a different granularity but enlarge the shifting step size to let each region be isolated from each other. 4) The proposed regional mixing.

\begin{figure}[ht]
    \centering
    \includegraphics[width=.5\columnwidth]{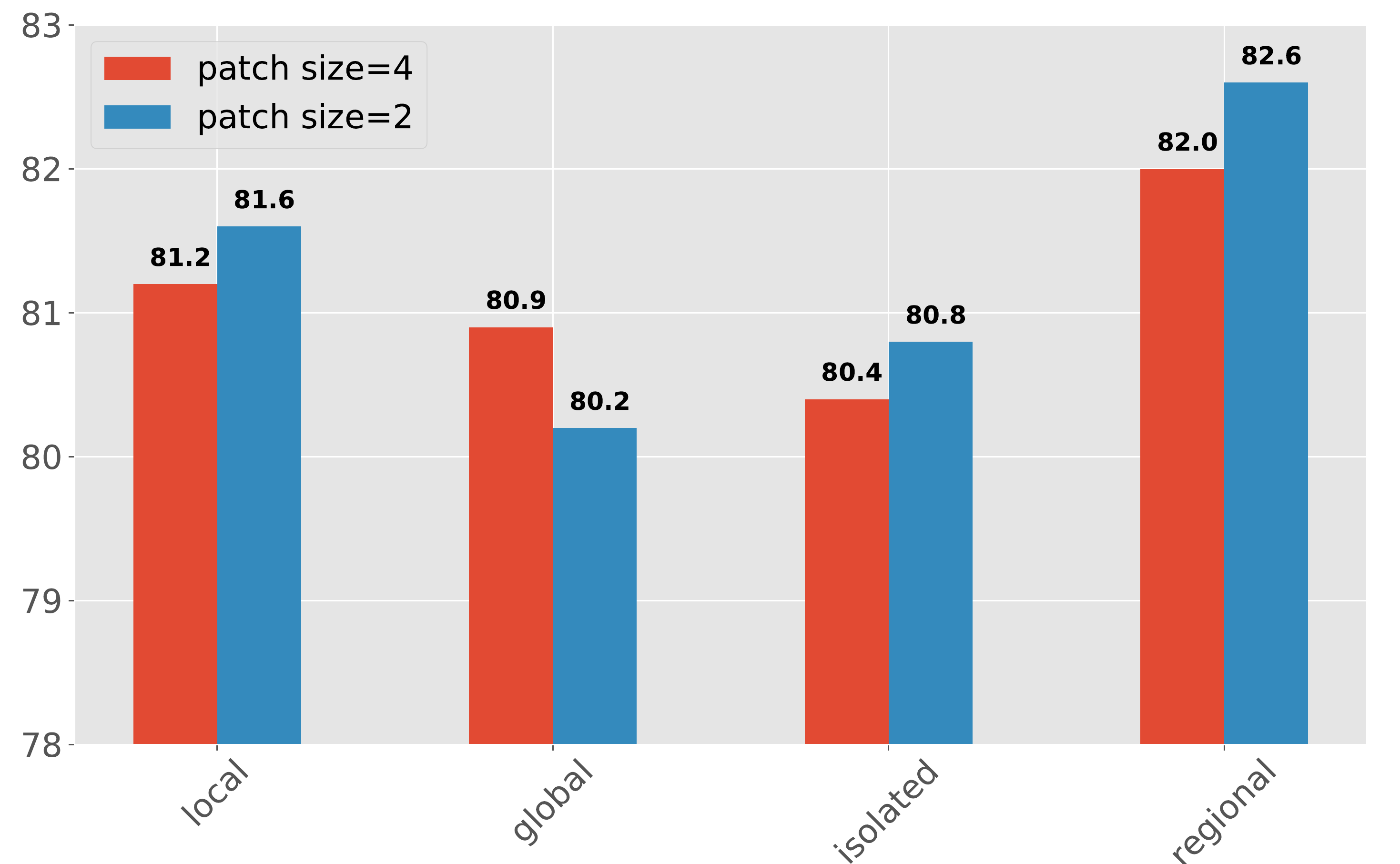}
    \vspace{-5mm}
    \caption{Image classification accuracy comparison with four configurations, as described in Table~\ref{tab-ablation}. Experiments are conducted on inputs of patch size 4 (red bins) and size 2 (blue bins).}
    \label{fig:ablation}
\end{figure}

We show the classification results with inputs that have a patch size of 2 or 4 in Fig.~\ref{fig:ablation}. As we can see, only using local or global mixing has a lower performance than regional mixing. Especially, for global mixing without the self-attention mechanism, token-mixing becomes less flexible to capture important information. We observe the performance is even lower when the patch size is ~2. For the isolated setting, without interactions within regions, information from different regions is hard to collect. This isolated mixing is even less effective than only using local information. When the patch size is 2, the number of input tokens becomes larger. We can see local and regional mixings both leverage finer-grained information improving classification accuracy, with significant improvements in regional mixing. This continues to validate the effectiveness of the regional mixing.

When combining MS-MLP with Transformers, we inject MS-MLP in both stages 0 and 1 of the Transformer to match the model parameters and throughputs. We study whether the Transformer itself can benefit from smaller input patches. We show a comparison between Swin-T and MS-MLP+Swin-T in Table~\ref{tab-patch}, where we can see smaller patch inputs only improve Swin-T by 0.1\%. However, the improvements from MS-MLP are more significant for both patch size 2 (81.4\% vs. 81.9\%) and 4 (81.3\% vs. 81.6\%).

\section{Conclusion}
In this paper, we present a regional mixing method via mixing and shifting operations to enable an efficient modeling of local-global dependency in MLPs, MS-MLP performs the token mixing at both fine-grain and coarse-grain levels, effectively handling both local and global dependencies with low computational cost. Extensive experiments show the effectiveness of MS-MLP over SoTA MLPs and other representative SoTA methods on  both image classification and object detection and segmentation. 

\section*{Acknowledgement}
The authors want to thank Davis Mueller for proofreading and providing insightful comments of this paper.

\bibliography{ref}

\begin{thebibliography}{57}
\providecommand{\natexlab}[1]{#1}
\providecommand{\url}[1]{\texttt{#1}}
\expandafter\ifx\csname urlstyle\endcsname\relax
  \providecommand{\doi}[1]{doi: #1}\else
  \providecommand{\doi}{doi: \begingroup \urlstyle{rm}\Url}\fi

\bibitem[Ba et~al.(2016)Ba, Kiros, and Hinton]{ln}
Ba, J.~L., Kiros, J.~R., and Hinton, G.~E.
\newblock Layer normalization.
\newblock \emph{arXiv preprint arXiv:1607.06450}, 2016.

\bibitem[Cai \& Vasconcelos(2018)Cai and Vasconcelos]{cascadercnn}
Cai, Z. and Vasconcelos, N.
\newblock Cascade r-cnn: Delving into high quality object detection.
\newblock In \emph{Proceedings of the IEEE conference on computer vision and
  pattern recognition}, 2018.

\bibitem[Chen et~al.(2021{\natexlab{a}})Chen, Li, Li, Li, Bai, Lin, Sun, Yan,
  and Ouyang]{chen2021glit}
Chen, B., Li, P., Li, C., Li, B., Bai, L., Lin, C., Sun, M., Yan, J., and
  Ouyang, W.
\newblock Glit: Neural architecture search for global and local image
  transformer.
\newblock In \emph{Proceedings of the IEEE/CVF International Conference on
  Computer Vision}, pp.\  12--21, 2021{\natexlab{a}}.

\bibitem[Chen et~al.(2019)Chen, Wang, Pang, Cao, Xiong, Li, Sun, Feng, Liu, Xu,
  Zhang, Cheng, Zhu, Cheng, Zhao, Li, Lu, Zhu, Wu, Dai, Wang, Shi, Ouyang, Loy,
  and Lin]{mmdetection}
Chen, K., Wang, J., Pang, J., Cao, Y., Xiong, Y., Li, X., Sun, S., Feng, W.,
  Liu, Z., Xu, J., Zhang, Z., Cheng, D., Zhu, C., Cheng, T., Zhao, Q., Li, B.,
  Lu, X., Zhu, R., Wu, Y., Dai, J., Wang, J., Shi, J., Ouyang, W., Loy, C.~C.,
  and Lin, D.
\newblock {MMDetection}: Open mmlab detection toolbox and benchmark.
\newblock \emph{arXiv preprint arXiv:1906.07155}, 2019.

\bibitem[Chen et~al.(2021{\natexlab{b}})Chen, Xie, Ge, Liang, and
  Luo]{chen2021cyclemlp}
Chen, S., Xie, E., Ge, C., Liang, D., and Luo, P.
\newblock Cyclemlp: A mlp-like architecture for dense prediction.
\newblock \emph{arXiv preprint arXiv:2107.10224}, 2021{\natexlab{b}}.

\bibitem[Contributors(2020)]{mmseg}
Contributors, M.
\newblock {MMSegmentation}: Openmmlab semantic segmentation toolbox and
  benchmark.
\newblock \url{https://github.com/open-mmlab/mmsegmentation}, 2020.

\bibitem[Cubuk et~al.(2020)Cubuk, Zoph, Shlens, and Le]{randaugment}
Cubuk, E.~D., Zoph, B., Shlens, J., and Le, Q.~V.
\newblock Randaugment: Practical automated data augmentation with a reduced
  search space.
\newblock In \emph{Proceedings of the IEEE/CVF Conference on Computer Vision
  and Pattern Recognition Workshops}, 2020.

\bibitem[Deng et~al.(2009)Deng, Dong, Socher, Li, Li, and
  Fei-Fei]{deng2009imagenet}
Deng, J., Dong, W., Socher, R., Li, L.-J., Li, K., and Fei-Fei, L.
\newblock Imagenet: A large-scale hierarchical image database.
\newblock In \emph{2009 IEEE conference on computer vision and pattern
  recognition}, pp.\  248--255. Ieee, 2009.

\bibitem[Dosovitskiy et~al.(2020)Dosovitskiy, Beyer, Kolesnikov, Weissenborn,
  Zhai, Unterthiner, Dehghani, Minderer, Heigold, Gelly, et~al.]{vit}
Dosovitskiy, A., Beyer, L., Kolesnikov, A., Weissenborn, D., Zhai, X.,
  Unterthiner, T., Dehghani, M., Minderer, M., Heigold, G., Gelly, S., et~al.
\newblock An image is worth 16x16 words: Transformers for image recognition at
  scale.
\newblock \emph{arXiv preprint arXiv:2010.11929}, 2020.

\bibitem[Guo et~al.(2021)Guo, Tang, Han, Chen, Wu, Xu, Xu, and
  Wang]{guo2021hire}
Guo, J., Tang, Y., Han, K., Chen, X., Wu, H., Xu, C., Xu, C., and Wang, Y.
\newblock Hire-mlp: Vision mlp via hierarchical rearrangement.
\newblock \emph{arXiv preprint arXiv:2108.13341}, 2021.

\bibitem[Han et~al.(2021)Han, Xiao, Wu, Guo, Xu, and Wang]{han2021transformer}
Han, K., Xiao, A., Wu, E., Guo, J., Xu, C., and Wang, Y.
\newblock Transformer in transformer.
\newblock \emph{arXiv preprint arXiv:2103.00112}, 2021.

\bibitem[He et~al.(2016{\natexlab{a}})He, Zhang, Ren, and Sun]{he2016deep}
He, K., Zhang, X., Ren, S., and Sun, J.
\newblock Deep residual learning for image recognition.
\newblock In \emph{Proceedings of the IEEE conference on computer vision and
  pattern recognition}, pp.\  770--778, 2016{\natexlab{a}}.

\bibitem[He et~al.(2016{\natexlab{b}})He, Zhang, Ren, and Sun]{resnet}
He, K., Zhang, X., Ren, S., and Sun, J.
\newblock Deep residual learning for image recognition.
\newblock In \emph{Proceedings of the IEEE conference on computer vision and
  pattern recognition}, pp.\  770--778, 2016{\natexlab{b}}.

\bibitem[He et~al.(2017)He, Gkioxari, Doll{\'a}r, and Girshick]{maskrcnn}
He, K., Gkioxari, G., Doll{\'a}r, P., and Girshick, R.
\newblock Mask r-cnn.
\newblock In \emph{Proceedings of the IEEE international conference on computer
  vision}, 2017.

\bibitem[Hou et~al.(2021)Hou, Jiang, Yuan, Cheng, Yan, and Feng]{hou2021vision}
Hou, Q., Jiang, Z., Yuan, L., Cheng, M.-M., Yan, S., and Feng, J.
\newblock Vision permutator: A permutable mlp-like architecture for visual
  recognition.
\newblock \emph{arXiv preprint arXiv:2106.12368}, 2021.

\bibitem[Huang et~al.(2016)Huang, Sun, Liu, Sedra, and Weinberger]{droppath}
Huang, G., Sun, Y., Liu, Z., Sedra, D., and Weinberger, K.~Q.
\newblock Deep networks with stochastic depth.
\newblock In \emph{European conference on computer vision}, 2016.

\bibitem[Ioffe \& Szegedy(2015)Ioffe and Szegedy]{bn}
Ioffe, S. and Szegedy, C.
\newblock Batch normalization: Accelerating deep network training by reducing
  internal covariate shift.
\newblock In \emph{International conference on machine learning}, 2015.

\bibitem[Krizhevsky et~al.(2012)Krizhevsky, Sutskever, and Hinton]{alexnet}
Krizhevsky, A., Sutskever, I., and Hinton, G.~E.
\newblock Imagenet classification with deep convolutional neural networks.
\newblock \emph{Advances in neural information processing systems}, pp.\
  1097--1105, 2012.

\bibitem[LeCun et~al.(1995)LeCun, Bengio, et~al.]{lecun1995convolutional}
LeCun, Y., Bengio, Y., et~al.
\newblock Convolutional networks for images, speech, and time series.
\newblock \emph{The handbook of brain theory and neural networks},
  3361\penalty0 (10):\penalty0 1995, 1995.

\bibitem[LeCun et~al.(1998)LeCun, Bottou, Bengio, and
  Haffner]{lecun1998gradient}
LeCun, Y., Bottou, L., Bengio, Y., and Haffner, P.
\newblock Gradient-based learning applied to document recognition.
\newblock \emph{Proceedings of the IEEE}, 86\penalty0 (11):\penalty0
  2278--2324, 1998.

\bibitem[Li et~al.(2021)Li, Hassani, Walton, and Shi]{li2021convmlp}
Li, J., Hassani, A., Walton, S., and Shi, H.
\newblock Convmlp: Hierarchical convolutional mlps for vision.
\newblock \emph{arXiv preprint arXiv:2109.04454}, 2021.

\bibitem[Lian et~al.(2021)Lian, Yu, Sun, and Gao]{lian2021mlp}
Lian, D., Yu, Z., Sun, X., and Gao, S.
\newblock As-mlp: An axial shifted mlp architecture for vision.
\newblock \emph{arXiv preprint arXiv:2107.08391}, 2021.

\bibitem[Lin et~al.(2014)Lin, Maire, Belongie, Hays, Perona, Ramanan,
  Doll{\'a}r, and Zitnick]{coco}
Lin, T.-Y., Maire, M., Belongie, S., Hays, J., Perona, P., Ramanan, D.,
  Doll{\'a}r, P., and Zitnick, C.~L.
\newblock Microsoft coco: Common objects in context.
\newblock In \emph{European conference on computer vision}, 2014.

\bibitem[Lin et~al.(2017)Lin, Goyal, Girshick, He, and Doll{\'a}r]{retinanet}
Lin, T.-Y., Goyal, P., Girshick, R., He, K., and Doll{\'a}r, P.
\newblock Focal loss for dense object detection.
\newblock In \emph{Proceedings of the IEEE international conference on computer
  vision}, 2017.

\bibitem[Liu et~al.(2021{\natexlab{a}})Liu, Dai, So, and Le]{gmlp}
Liu, H., Dai, Z., So, D.~R., and Le, Q.~V.
\newblock Pay attention to mlps.
\newblock \emph{arXiv preprint arXiv:2105.08050}, 2021{\natexlab{a}}.

\bibitem[Liu et~al.(2021{\natexlab{b}})Liu, Lin, Cao, Hu, Wei, Zhang, Lin, and
  Guo]{liu2021Swin}
Liu, Z., Lin, Y., Cao, Y., Hu, H., Wei, Y., Zhang, Z., Lin, S., and Guo, B.
\newblock Swin transformer: Hierarchical vision transformer using shifted
  windows.
\newblock \emph{arXiv preprint arXiv:2103.14030}, 2021{\natexlab{b}}.

\bibitem[Liu et~al.(2022)Liu, Mao, Wu, Feichtenhofer, Darrell, and
  Xie]{liu2022convnet}
Liu, Z., Mao, H., Wu, C.-Y., Feichtenhofer, C., Darrell, T., and Xie, S.
\newblock A convnet for the 2020s.
\newblock \emph{arXiv preprint arXiv:2201.03545}, 2022.

\bibitem[Loshchilov \& Hutter(2017)Loshchilov and Hutter]{adamw}
Loshchilov, I. and Hutter, F.
\newblock Decoupled weight decay regularization.
\newblock \emph{arXiv preprint arXiv:1711.05101}, 2017.

\bibitem[Lou et~al.(2021)Lou, Xue, Zheng, and You]{lou2021sparse}
Lou, Y., Xue, F., Zheng, Z., and You, Y.
\newblock Sparse-mlp: A fully-mlp architecture with conditional computation.
\newblock \emph{arXiv preprint arXiv:2109.02008}, 2021.

\bibitem[Mao et~al.(2021)Mao, Qi, Chen, Li, Ye, He, and Xue]{mao2021rethinking}
Mao, X., Qi, G., Chen, Y., Li, X., Ye, S., He, Y., and Xue, H.
\newblock Rethinking the design principles of robust vision transformer.
\newblock \emph{arXiv preprint arXiv:2105.07926}, 2021.

\bibitem[Melas-Kyriazi(2021)]{melas2021you}
Melas-Kyriazi, L.
\newblock Do you even need attention? a stack of feed-forward layers does
  surprisingly well on imagenet.
\newblock \emph{arXiv preprint arXiv:2105.02723}, 2021.

\bibitem[Polyak \& Juditsky(1992)Polyak and Juditsky]{ema}
Polyak, B.~T. and Juditsky, A.~B.
\newblock Acceleration of stochastic approximation by averaging.
\newblock \emph{SIAM journal on control and optimization}, 30\penalty0
  (4):\penalty0 838--855, 1992.

\bibitem[Radosavovic et~al.(2020)Radosavovic, Kosaraju, Girshick, He, and
  Doll{\'a}r]{radosavovic2020designing}
Radosavovic, I., Kosaraju, R.~P., Girshick, R., He, K., and Doll{\'a}r, P.
\newblock Designing network design spaces.
\newblock In \emph{Proceedings of the IEEE/CVF Conference on Computer Vision
  and Pattern Recognition}, pp.\  10428--10436, 2020.

\bibitem[Rao et~al.(2021)Rao, Zhao, Zhu, Lu, and Zhou]{rao2021global}
Rao, Y., Zhao, W., Zhu, Z., Lu, J., and Zhou, J.
\newblock Global filter networks for image classification.
\newblock \emph{arXiv preprint arXiv:2107.00645}, 2021.

\bibitem[Simonyan \& Zisserman(2014)Simonyan and Zisserman]{vggnet}
Simonyan, K. and Zisserman, A.
\newblock Very deep convolutional networks for large-scale image recognition.
\newblock \emph{arXiv preprint arXiv:1409.1556}, 2014.

\bibitem[Szegedy et~al.(2015)Szegedy, Liu, Jia, Sermanet, Reed, Anguelov,
  Erhan, Vanhoucke, and Rabinovich]{googlenet}
Szegedy, C., Liu, W., Jia, Y., Sermanet, P., Reed, S., Anguelov, D., Erhan, D.,
  Vanhoucke, V., and Rabinovich, A.
\newblock Going deeper with convolutions.
\newblock In \emph{Proceedings of the IEEE conference on computer vision and
  pattern recognition}, pp.\  1--9, 2015.

\bibitem[Szegedy et~al.(2016)Szegedy, Vanhoucke, Ioffe, Shlens, and
  Wojna]{labelsmooth}
Szegedy, C., Vanhoucke, V., Ioffe, S., Shlens, J., and Wojna, Z.
\newblock Rethinking the inception architecture for computer vision.
\newblock In \emph{Proceedings of the IEEE conference on computer vision and
  pattern recognition}, 2016.

\bibitem[Tang et~al.(2021{\natexlab{a}})Tang, Zhao, Wang, Luo, Xie, and
  Zeng]{tang2021sparse}
Tang, C., Zhao, Y., Wang, G., Luo, C., Xie, W., and Zeng, W.
\newblock Sparse mlp for image recognition: Is self-attention really necessary?
\newblock \emph{arXiv preprint arXiv:2109.05422}, 2021{\natexlab{a}}.

\bibitem[Tang et~al.(2021{\natexlab{b}})Tang, Han, Guo, Xu, Li, Xu, and
  Wang]{tang2021image}
Tang, Y., Han, K., Guo, J., Xu, C., Li, Y., Xu, C., and Wang, Y.
\newblock An image patch is a wave: Phase-aware vision mlp.
\newblock \emph{arXiv preprint arXiv:2111.12294}, 2021{\natexlab{b}}.

\bibitem[Tatsunami \& Taki(2021)Tatsunami and Taki]{tatsunami2021raftmlp}
Tatsunami, Y. and Taki, M.
\newblock Raftmlp: Do mlp-based models dream of winning over computer vision?
\newblock \emph{arXiv preprint arXiv:2108.04384}, 2021.

\bibitem[Tolstikhin et~al.(2021)Tolstikhin, Houlsby, Kolesnikov, Beyer, Zhai,
  Unterthiner, Yung, Keysers, Uszkoreit, Lucic, et~al.]{mlp-mixer}
Tolstikhin, I., Houlsby, N., Kolesnikov, A., Beyer, L., Zhai, X., Unterthiner,
  T., Yung, J., Keysers, D., Uszkoreit, J., Lucic, M., et~al.
\newblock Mlp-mixer: An all-mlp architecture for vision.
\newblock \emph{arXiv preprint arXiv:2105.01601}, 2021.

\bibitem[Touvron et~al.(2020)Touvron, Cord, Douze, Massa, Sablayrolles, and
  J{\'e}gou]{touvron2020training}
Touvron, H., Cord, M., Douze, M., Massa, F., Sablayrolles, A., and J{\'e}gou,
  H.
\newblock Training data-efficient image transformers \& distillation through
  attention.
\newblock \emph{arXiv preprint arXiv:2012.12877}, 2020.

\bibitem[Touvron et~al.(2021)Touvron, Bojanowski, Caron, Cord, El-Nouby, Grave,
  Joulin, Synnaeve, Verbeek, and J{\'e}gou]{resmlp}
Touvron, H., Bojanowski, P., Caron, M., Cord, M., El-Nouby, A., Grave, E.,
  Joulin, A., Synnaeve, G., Verbeek, J., and J{\'e}gou, H.
\newblock Resmlp: Feedforward networks for image classification with
  data-efficient training.
\newblock \emph{arXiv preprint arXiv:2105.03404}, 2021.

\bibitem[Vaswani et~al.(2017)Vaswani, Shazeer, Parmar, Uszkoreit, Jones, Gomez,
  Kaiser, and Polosukhin]{vaswani2017attention}
Vaswani, A., Shazeer, N., Parmar, N., Uszkoreit, J., Jones, L., Gomez, A.~N.,
  Kaiser, {\L}., and Polosukhin, I.
\newblock Attention is all you need.
\newblock In \emph{Advances in neural information processing systems}, pp.\
  5998--6008, 2017.

\bibitem[Wang et~al.(2021{\natexlab{a}})Wang, Xie, Li, Fan, Song, Liang, Lu,
  Luo, and Shao]{pvt}
Wang, W., Xie, E., Li, X., Fan, D.-P., Song, K., Liang, D., Lu, T., Luo, P.,
  and Shao, L.
\newblock Pyramid vision transformer: A versatile backbone for dense prediction
  without convolutions.
\newblock \emph{arXiv preprint arXiv:2102.12122}, 2021{\natexlab{a}}.

\bibitem[Wang et~al.(2021{\natexlab{b}})Wang, Xie, Li, Fan, Song, Liang, Lu,
  Luo, and Shao]{wang2021pyramid}
Wang, W., Xie, E., Li, X., Fan, D.-P., Song, K., Liang, D., Lu, T., Luo, P.,
  and Shao, L.
\newblock Pyramid vision transformer: A versatile backbone for dense prediction
  without convolutions.
\newblock \emph{arXiv preprint arXiv:2102.12122}, 2021{\natexlab{b}}.

\bibitem[Wu et~al.(2021{\natexlab{a}})Wu, Xiao, Codella, Liu, Dai, Yuan, and
  Zhang]{wu2021cvt}
Wu, H., Xiao, B., Codella, N., Liu, M., Dai, X., Yuan, L., and Zhang, L.
\newblock Cvt: Introducing convolutions to vision transformers.
\newblock \emph{arXiv preprint arXiv:2103.15808}, 2021{\natexlab{a}}.

\bibitem[Wu et~al.(2021{\natexlab{b}})Wu, Peng, Chen, Fu, and
  Chao]{wu2021rethinking}
Wu, K., Peng, H., Chen, M., Fu, J., and Chao, H.
\newblock Rethinking and improving relative position encoding for vision
  transformer.
\newblock In \emph{Proceedings of the IEEE/CVF International Conference on
  Computer Vision}, pp.\  10033--10041, 2021{\natexlab{b}}.

\bibitem[Xiao et~al.(2018)Xiao, Liu, Zhou, Jiang, and Sun]{upernet}
Xiao, T., Liu, Y., Zhou, B., Jiang, Y., and Sun, J.
\newblock Unified perceptual parsing for scene understanding.
\newblock In \emph{Proceedings of the European Conference on Computer Vision
  (ECCV)}, 2018.

\bibitem[Yang et~al.(2021)Yang, Li, Zhang, Dai, Xiao, Yuan, and
  Gao]{yang2021focal}
Yang, J., Li, C., Zhang, P., Dai, X., Xiao, B., Yuan, L., and Gao, J.
\newblock Focal attention for long-range interactions in vision transformers.
\newblock \emph{Advances in Neural Information Processing Systems}, 34, 2021.

\bibitem[Yu et~al.(2021)Yu, Li, Cai, Sun, and Li]{yu2021s}
Yu, T., Li, X., Cai, Y., Sun, M., and Li, P.
\newblock S$^2$-mlp: Spatial-shift mlp architecture for vision.
\newblock \emph{arXiv preprint arXiv:2106.07477}, 2021.

\bibitem[Yu et~al.(2022)Yu, Li, Cai, Sun, and Li]{yu2022s2}
Yu, T., Li, X., Cai, Y., Sun, M., and Li, P.
\newblock S$^2$-mlp: Spatial-shift mlp architecture for vision.
\newblock In \emph{Proceedings of the IEEE/CVF Winter Conference on
  Applications of Computer Vision}, pp.\  297--306, 2022.

\bibitem[Yuan et~al.(2021)Yuan, Chen, Wang, Yu, Shi, Jiang, Tay, Feng, and
  Yan]{yuan2021tokens}
Yuan, L., Chen, Y., Wang, T., Yu, W., Shi, Y., Jiang, Z., Tay, F.~E., Feng, J.,
  and Yan, S.
\newblock Tokens-to-token vit: Training vision transformers from scratch on
  imagenet.
\newblock \emph{arXiv preprint arXiv:2101.11986}, 2021.

\bibitem[Yun et~al.(2019)Yun, Han, Oh, Chun, Choe, and Yoo]{cutmix}
Yun, S., Han, D., Oh, S.~J., Chun, S., Choe, J., and Yoo, Y.
\newblock Cutmix: Regularization strategy to train strong classifiers with
  localizable features.
\newblock In \emph{Proceedings of the IEEE/CVF International Conference on
  Computer Vision}, 2019.

\bibitem[Zhang et~al.(2017)Zhang, Cisse, Dauphin, and Lopez-Paz]{mixup}
Zhang, H., Cisse, M., Dauphin, Y.~N., and Lopez-Paz, D.
\newblock mixup: Beyond empirical risk minimization.
\newblock \emph{arXiv preprint arXiv:1710.09412}, 2017.

\bibitem[Zhong et~al.(2020)Zhong, Zheng, Kang, Li, and Yang]{randomerasing}
Zhong, Z., Zheng, L., Kang, G., Li, S., and Yang, Y.
\newblock Random erasing data augmentation.
\newblock In \emph{Proceedings of the AAAI Conference on Artificial
  Intelligence}, 2020.

\bibitem[Zhou et~al.(2019)Zhou, Zhao, Puig, Xiao, Fidler, Barriuso, and
  Torralba]{ade20k}
Zhou, B., Zhao, H., Puig, X., Xiao, T., Fidler, S., Barriuso, A., and Torralba,
  A.
\newblock Semantic understanding of scenes through the ade20k dataset.
\newblock \emph{International Journal of Computer Vision}, 2019.

\end{thebibliography}
\bibliographystyle{icml2022}

\newpage
\appendix
\onecolumn
\section{Experiment settings}\label{app:exp-settings}
\subsection{Image classification on ImageNet-1K}
Following the settings of the Swin transformer~\citep{liu2021Swin}, image classification is performed by applying an adaptive global average pooling layer on the output feature map of the last stage followed by a linear classifier. In evaluation, the top-1 accuracy using a single crop is reported. 

The training settings mostly follow the Swin transformer~\citep{liu2021Swin} and AS-MLP~\citep{lian2021mlp}. For all model variants, we adopt a default input image resolution of $224 \times 224$. When training from scratch with an input of size $224 \times 224$, we employ an AdamW~\citep{adamw} optimizer for 300 epochs using a cosine decay learning rate scheduler with 20 epochs of linear warm-up. Keeping a batch size of 1,024, an initial learning rate of
0.001, a weight decay of 0.05, and a gradient clipping with a max norm of 1 are used. We include most of the augmentation and regularization strategies of \citep{touvron2020training} in training, including Rand-Augment~\citep{randaugment}, MixUp~\citep{mixup}, CutMix~\citep{cutmix}, Label Smoothing~\citep{labelsmooth}, Random Erasing~\citep{randomerasing}, and DropPath~\citep{droppath}. Different from the Swin and Focal-Attention Transformer, we empirically find the Exponential Moving Average (EMA)~\citep{ema} can enhance performance, but for a fair comparison we did not report the EMA results in the paper. An increasing degree of stochastic depth augmentation is employed for larger models, \textit{i.e.}, 0.2, 0.3, 0.5 for MLP-T, MLP-S, and MLP-B, respectively.

\subsection{Downstream tasks: Object detection on COCO and semantic segmentation on ADE20K}
On COCO-2017 tasks, we consider four typical object detection frameworks: RetinaNet~\citep{retinanet}, Mask R-CNN~\citep{maskrcnn}, and Cascade Mask R-CNN~\citep{cascadercnn} in mmdetection~\citep{mmdetection}. We utilize the single-scale training and multi-scale training for the ``1x" and ``3x" (resizing the input such that the shorter side is between 480 and 800 while the longer side is at most 1,333) schedules, respectively. For these frameworks, we utilize the same settings where we adopt the AdamW optimizer~\citep{adamw} (initial learning rate of 0.0001, weight decay of 0.05, and a batch size of 16), and ``3x" schedule (36 epochs with the learning rate decayed by 10× at epochs 27 and 33).

On ADE20K~\citep{ade20k}, which contains 20,210 training images and 2,000 validation images, we conduct semantic segmentation experiments following the settings in \citep{pvt,chen2021cyclemlp,liu2021Swin}. We use UperNet~\citep{upernet} as the segmentation framework and a pretrained MS-MLP as the backbone. For all models, we use a standard recipe by setting the input size to $512 \times 512$ and train the model for 160k iterations with a batch size 16 in the training. We follow the training settings described in~\citet{liu2021Swin}, where we employ the AdamW optimizer~\citep{adamw} with an initial learning rate of $6 \times 10^{-5}$, a weight decay of 0.01, a scheduler that uses linear learning rate decay, and a linear warmup of 1,500 iterations. 
For augmentations, we adopt the default setting in mmsegmentation~\citep{mmseg} of random horizontal flipping, random re-scaling within ratio range $[0.5, 2.0]$, and random photometric distortion. Stochastic depth with ratio of 0.2 is applied for all models.

\section{Additional Experiment Results.}
\subsection{Instance segmentation with 3x schedule}
For the object detection and instance segmentation tasks on COCO-2017, we also train models with 3× schedule and multi-scale training strategy as described in the main experiments. The results of Mask R-CNN~\citep{maskrcnn} and Cascade Mask R-CNN~\citep{cascadercnn} are shown in Table~\ref{table:coco-3x}. Similar to the results in the ``1x" schedule, the proposed MS-MLP achieves a higher performance.

\begin{table*}[t]
\centering
\caption{Instance segmentation results on COCO val2017. Mask R-CNN and Cascade Mask R-CNN are trained on the ``3x" schedule.}
\label{table:coco-3x}
\renewcommand{\arraystretch}{0.88}
\setlength\tabcolsep{4.2pt}
\resizebox{\textwidth}{!}{\begin{tabular}{l|c|ccc|ccc|c|ccc|ccc}
\toprule[1.5pt]
\multirow{2}{*}{Backbone} &\multicolumn{7}{c|}{Mask R-CNN 3$\times$} &\multicolumn{7}{c}{Cascade Mask R-CNN 3$\times$} \\
\cline{2-15} 
& FLOPs & AP$^{\rm b}$ & AP$_{50}^{\rm b}$ & AP$_{75}^{\rm b}$ & AP$^{\rm m}$ & AP$_{50}^{\rm m}$ & AP$_{75}^{\rm m}$ & FLOPs & AP$^{\rm b}$ & AP$_{50}^{\rm b}$ &AP$_{75}^{\rm b}$  &AP$^{\rm m}$ &AP$_{50}^{\rm m}$ & AP$_{75}^{\rm m}$\\
\midrule  

ResNet50~\citep{resnet} & 260G & 41.0 & 61.7 & 44.9 & 37.1 & 58.4 & 40.1 & 738G & 46.3 & 64.3 & 50.5 & 40.1 & 61.7 & 43.4 \\
AS-MLP-T~\citep{lian2021mlp} & 260G & 46.0 & 67.5 & 50.7 & 41.5 & 64.6 & 44.5 & 739G & 50.1 & 68.8 & 54.3 & 43.5 & 66.3 & 46.9 \\
Swin-T~\citep{liu2021Swin} & 264G & 46.0 & {68.2} & 50.2 & 41.6 & 65.1 & 44.8 & 742G & 50.5 & 69.3 & 54.9 & 43.7 & 66.6 & 47.1 \\
Hire-MLP-Small~\citep{guo2021hire} & 256G & {46.2} & {68.2} & {50.9} & {42.0} & {65.6} & {45.3} & 734G & {50.7} & {69.4} & {55.1} & {44.2} & {66.9} & {48.1} \\
\rowcolor{Gray}
MS-MLP-T (ours) & 262G & 46.2 & 67.8 & 50.8 & 41.7 & 65.2 & 45.0 & 744G & 50.4 & 69.2 & 54.6 & 43.7 & 66.5 & 47.6 \\
\midrule

Swin-S~\citep{liu2021Swin} & 354G & {48.5} & {70.2} & {53.5} & {43.3} & {67.3} & 46.6 & 832G & {51.8} & {70.4} & {56.3} & 44.7 & {67.9} & {48.5} \\
AS-MLP-S~\citep{lian2021mlp} & 346G & 47.8 & 68.9 & 52.5 & 42.9 & 66.4 & 46.3 & 823G & 51.1 & 69.8 & 55.6 & 44.2 & 67.3 & 48.1 \\
Hire-MLP-Base~\citep{guo2021hire} & 334G & 48.1 & 69.6 & 52.7 & 43.1 & 66.8 & {46.7} & 813G & 51.7 & 70.2 & 56.1 & {44.8} & 67.8 & {48.5} \\
\rowcolor{Gray}
MS-MLP-S (ours) & 424G & 48.6 & 70.8 & 53.4 & 43.7 & 67.7 & 47.2 & 841G & 51.9 & 70.8 & 56.6 & 43.7 & 66.5 & 47.3 \\\midrule
Swin-B~\citep{liu2021Swin} & 496G & {48.5} & {70.2} & {53.5} & {43.3} & {67.3} & 46.6 & 982G & {51.9} & {71.8} & {57.5} & 45.8 & {67.4} & {49.7} \\
\rowcolor{Gray}
MS-MLP-B (ours) & 561G & 49.0 & 70.0 & 52.6 & 43.7 & 65.4 & 46.7 & 986G & 52.6 & 71.4 & 57.2 & 45.4 & 68.9 & 49.3 \\
   
\bottomrule[1.5pt]
\end{tabular}}
\end{table*}

\subsection{Image classification with different architecture}
In the main paper we set the MS-MLP in an architecture with four-stages, where the number of blocks keeps a ratio of 1:1:3:1. To match the parameter size and throughputs, these stages contain 3-3-9-3 blocks for MS-MLP-T and 3-3-27-3 for MS-MLP-S and MS-MLP-B. We also match the architecture with Swin~\citep{liu2021Swin} and AS-MLP~\citep{lian2021mlp}, using the stage design of 2-2-6-2 for MS-MLP-T and 2-2-18-2 for MS-MLP-S and MS-MLP-B. The results are summarized in Table~\ref{tab-stage}. We can see the results are comparable to the results shown for Swin~\citep{liu2021Swin} and AS-MLP~\citep{lian2021mlp}, while the parameter size and FLOPs are fewer.

For the regional mixing, we choose to use depth-wise convolution with different kernel size. We conduct experiments to see the effects of the kernel sizes and convolution type as an additional ablation. In Table~\ref{tab-mixing}, we can observe the full convolution has no special effects to the final results, while using full convolution increases the FLOPs and slows down the training. We also observe slightly changing the kernel size does not affect the results. However, the results show that decreasing the region size while the relative distance increases degrades model performance.

\begin{table}[ht]
  \begin{minipage}[t]{.58\textwidth}
  \centering
    \caption{Image classification results of different architecture MS-MLP.} 
	\label{tab-stage}
    \renewcommand{\arraystretch}{1.2}
    \setlength{\tabcolsep}{1.0mm}{ 
	\setlength\tabcolsep{2pt}
    \resizebox{.95\columnwidth}{!}{\begin{tabular}{l|c|c|c|c}
    \toprule[1.5pt]
    Model & Blocks & \#Parameters & FLOPs & Top-1 acc.\\ \hline
    \multirow{2}{*}{MS-MLP-T}& 2-2-6-2 &  24M& 4.3G & 81.4\% \\ 
                               & 2-2-2-6-2 & 24M& 4.4G & 81.3\% \\ \hline
    \multirow{2}{*}{MS-MLP-S}& 2-2-18-2 & 42M & 7.8G & 82.8\% \\ 
                               & 2-2-2-18-2 & 42M & 7.8G & 83.0\% \\ \hline 
    \multirow{2}{*}{MS-MLP-B}& 2-2-18-2 & 74M & 13.8G & 83.3\% \\ 
                               & 2-2-2-18-2 & 74M& 13.9G & 83.2\% \\ \hline 
    \bottomrule[1.5pt]
    \end{tabular}
    }
    }
\end{minipage}\hfill
\begin{minipage}[t]{.4\textwidth}
  \centering
    \caption{Image classification results of different MS-block configurations.} 
	\label{tab-mixing}
    \renewcommand{\arraystretch}{1.}
    \setlength{\tabcolsep}{1.0mm}{ 
	\setlength\tabcolsep{2pt}
    \resizebox{.95\columnwidth}{!}{\begin{tabular}{l|c|c|c}
    \toprule[1.5pt]
     Region size & Conv type & FLOPs  & Top-1 acc.\\ \hline
     1-1-3-5-7 &  DW & 4.9G &  82.1\% \\ 
     1-1-3-5-7 & Full & 7.7G & 82.0\% \\ \hline
     1-3-5-7-9 & DW & 5.6G & 81.8\% \\ 
     1-3-5-7-9 & Full & 9.1G & 82.0\% \\ \hline 
     1-7-5-3-1 & DW & 4.9G &  81.1\% \\ 
     1-7-5-3-1 & Full & 7.7G & 81.3\% \\ \hline 
     1-5-3-3-1 & DW & 4.6G &  81.2\% \\ 
     1-5-3-3-1 & Full & 6.8G & 81.4\% \\ \hline 
    \bottomrule[1.5pt]
    \end{tabular}
    }
    }
\end{minipage}
\end{table}

\section{More discussion and limitations}
We demonstrate MS-MLP can perform as well as representative ViTs and CNNs on image classification, object detection, instance, and semantic segmentation tasks. While our goal is to offer a general way to handle global and local visual dependencies, MS-MLP still relies on a careful choice of the region size and the relative shifting distance, and we are not able to explore all possible configurations. As we realize computer vision applications are diverse, the current MS-MLP configuration may be suited for certain tasks, and we may need to explore a more general recipe for other tasks.

\begin{table}[t]
    \centering
    \caption{Full comparison of MS-MLP architecture with SOTA models on ImageNet-1K. }
	\label{tab-full-sota}
    \renewcommand{\arraystretch}{1.2}
    \setlength{\tabcolsep}{1.0mm}{ 
	\setlength\tabcolsep{2pt}
    \resizebox{.85\columnwidth}{!}{
    \begin{tabular}{l | c |c c c|l}
	\toprule[1.5pt]	
	
	\multirow{2}{*}{Model} & \multirow{2}{*}{Family}  & \multirow{2}{*}{Params.} & \multirow{2}{*}{FLOPs} & Throughput &  Top-1 \\ 
	&&&&(image / s)&acc. (\%)\\ \hline
	ResNet18~\citep{he2016deep}                   & CNN    & 12M & 1.8G  &-& 69.8 \\
	ResNet50~\citep{he2016deep}                   &  CNN   & 26M & 4.1G  &- &78.5 \\
	ResNet101~\citep{he2016deep}                  &  CNN  & 45M & 7.9G &-& 79.8 \\
	RegNetY-4G~\citep{radosavovic2020designing}   &  CNN   & 21M & 4.0G  &1157& 80.0 \\
	RegNetY-8G~\citep{radosavovic2020designing}   &  CNN  &  39M & 8.0G &592& 81.7 \\
	RegNetY-16G~\citep{radosavovic2020designing}   &  CNN  &  84M & 16.0G &335& 82.9 \\
	ConvNeXt-T~\citep{liu2022convnet}  & CNN & 29M & 4.5G & 774.7 & 82.1 \\
	ConvNeXt-S~\citep{liu2022convnet}  & CNN & 50M & 8.7G & 447.1 & 83.1 \\
	ConvNeXt-B~\citep{liu2022convnet}  & CNN & 89M & 15.4G & 292.1 & 83.8 \\
	\hline
	GFNet-H-S~\citep{rao2021global}               & FFT   &  32M & 4.5G  & -&81.5 \\
	GFNet-H-B~\citep{rao2021global}               & FFT   &  54M & 8.4G  & -&82.9 \\  \hline
	DeiT-S~\citep{touvron2020training}            & Trans  & 22M & 4.6G  & 940&79.8 \\
	DeiT-B~\citep{touvron2020training}            & Trans &  86M & 17.5G &  292&81.8 \\
	PVT-Small~\citep{wang2021pyramid}                 & Trans &  25M & 3.8G  &820& 79.8 \\
	PVT-Medium~\citep{wang2021pyramid}                 & Trans &  44M & 6.7G &526& 81.2 \\
	PVT-Large~\citep{wang2021pyramid}                 & Trans &  61M & 9.8G  &367& 81.7 \\
	T2T-ViT-14~\citep{yuan2021tokens} &Trans &22M&5.2G&764&81.5\\
	T2T-ViT-19~\citep{yuan2021tokens} &Trans &39M&8.9G&464&81.9\\
	T2T-ViT-24~\citep{yuan2021tokens} &Trans&64M&14.1G&312&82.3\\
	TNT-S~\citep{han2021transformer} & Trans  & 24M & 5.2G  &428& 81.5 \\
	TNT-B~\citep{han2021transformer}& Trans &  66M & 14.1G  &246& 82.9 \\
	iRPE-K~\citep{wu2021rethinking} &Trans & 87M&17.7G&- &82.4\\
	iRPE-QKV~\citep{wu2021rethinking}            & Trans  & 22M & 4.9G  &-& 81.4 \\

	GLiT-Small~\citep{chen2021glit}& Trans&25M&4.4G&-&80.5\\
	GLiT-Base~\citep{chen2021glit}& Trans&96M&17.0G&-&82.3\\ \hline
	\rowcolor{Gray}
    MS-MLP-T  (ours)     & MLP                       & 28M   & 4.9G  &792 &{82.1} \\
	\rowcolor{Gray}
    MS-MLP-S (ours)  & MLP & 50M   & 9.0G  &484 & {83.4}\\
	\rowcolor{Gray}
    MS-MLP-B      (ours)    &  MLP                         & 88M   & 16.1G  &366 & \textbf{83.8}\\ \hline
	Swin-T~\citep{liu2021Swin}                    & Trans &  29M & 4.5G  &755& 81.3 \\
	\rowcolor{Gray}
	MS-MLP + Swin-T   (ours)                  & MLP + T &  29M & 4.5G  &779 & \textbf{81.9}~\scriptsize{\textcolor{blue}{\bf(+0.6)}} \\
	Swin-S~\citep{liu2021Swin}                    & Trans & 50M & 8.7G  & 437&83.0 \\
	\rowcolor{Gray}
	MS-MLP  + Swin-S  (ours)                   & MLP + T & 50M & 8.7G  & 464 & \textbf{83.5}~\scriptsize{\textcolor{blue}{\bf(+0.5)}} \\	
	Swin-B~\citep{liu2021Swin}                    & Trans &  88M & 15.4G &278 &83.3 \\ 
	\rowcolor{Gray}
	MS-MLP  + Swin-B  (ours)                   & MLP + T &  88M & 15.4G &279 &\textbf{83.8}~\scriptsize{\textcolor{blue}{\bf(+0.5)}} \\ \hline
	Focal-Attention-T~\citep{yang2021focal}                    & Trans &  29M & 4.9G  &319& 82.2 \\
	\rowcolor{Gray}
	MS-MLP  + Focal-Attention-T    (ours)                 & MLP + T              &  29M & 5.6G  &451& \textbf{82.8}~\scriptsize{\textcolor{blue}{\bf(+0.6)}} \\
	Focal-Attention-S~\citep{yang2021focal}                    & Trans & 52M & 9.4G  & 192&83.5 \\
	\rowcolor{Gray}
	MS-MLP  + Focal-Attention-S    (ours)             & MLP + T                 & 52M & 10.1G  & 297 &\textbf{83.9}~\scriptsize{\textcolor{blue}{\bf(+0.4)}} \\	
	Focal-Attention-B~\citep{yang2021focal}                    & Trans &  90M & 16.4G &138 &83.8 \\ 
	\rowcolor{Gray}
	MS-MLP  + Focal-Attention-B    (ours)             & MLP + T                 & 90M & 17.6G & 207 &\textbf{84.0}~\scriptsize{\textcolor{blue}{\bf(+0.2)}} \\
	\bottomrule[1.5pt]
\end{tabular}}
    }\vspace{-3mm}
\end{table}


\end{document}